\documentclass[10pt,twocolumn,letterpaper]{article}

\usepackage[pagenumbers]{iccv} % To force page numbers, e.g. for an arXiv version

\PassOptionsToPackage{table,svgnames,dvipsnames,HTML}{xcolor}

\usepackage[table,svgnames,dvipsnames,HTML]{xcolor}
\usepackage{graphicx}
\usepackage{amsmath}
\usepackage{amssymb}
\usepackage{tabularx,booktabs}
\usepackage{colortbl}

\usepackage{bbding}
\usepackage{multirow}
\usepackage{enumitem}
\usepackage[most]{tcolorbox}

\newcolumntype{Y}{>{\centering\arraybackslash}X}
\definecolor{maroon}{RGB}{128,0,0}

\definecolor{iccvblue}{rgb}{0.21,0.49,0.74}
\usepackage[pagebackref,breaklinks,colorlinks,allcolors=iccvblue]{hyperref}

\def\paperID{770} % *** Enter the Paper ID here
\def\confName{ICCV}
\def\confYear{2025}

\title{Less is More: Empowering GUI Agent with Context-Aware Simplification}

\author{
Gongwei Chen$^{1}$, Xurui Zhou$^{1}$, Rui Shao$^{1}$\footnotemark[2], Yibo Lyu$^{1}$,
Kaiwen Zhou$^{2}$,\\
Shuai Wang$^{2}$, Wentao Li$^{2}$, Yinchuan Li$^{2}$, Zhongang Qi$^{2}$,
Liqiang Nie$^{1}$\footnotemark[2]
\\
    $^{1}$Harbin Institute of Technology, Shenzhen\quad
    $^{2}$Huawei Noah's Ark Lab\\
    \texttt{\normalsize{chengongwei@hit.edu.cn}}\hspace{0.5cm}\texttt{\normalsize{shaorui@hit.edu.cn}}\hspace{0.5cm}\texttt{\normalsize{nieliqiang@gmail.com}}\\
    \texttt{\normalsize{\url{https://github.com/JiuTian-VL/SimpAgent}}}
}

\begin{document}
\maketitle

\renewcommand{\thefootnote}{\fnsymbol{footnote}} 
\footnotetext[2]{Corresponding authors}

\begin{abstract}
    % Graphical User Interface (GUI) agents aim to autonomously accomplish tasks with the interface without manual intervention. 
    % The research focus of Graphical User Interface (GUI) agents is transitioning from text-dependent to pure-vision-based approaches.
    % While pure-vision-based GUI agents have been a promising and general choice, they primarily focus on the collection of comprehensive, high-quality pre-training data, ignoring the inherent challenges related to contextual modeling. 
    The research focus of GUI agents is shifting from text-dependent to pure-vision-based approaches, which, though promising, prioritize comprehensive pre-training data collection while neglecting contextual modeling challenges.
    We probe the characteristics of element and history contextual modeling in GUI agent and summarize: \textbf{1) }\textbf{the high-density and loose-relation of element context} highlight the existence of many unrelated elements and their negative influence;\textbf{ 2) }\textbf{the high redundancy of history context} reveals the inefficient history modeling in current GUI agents.
    In this work, we propose a context-aware simplification framework for building an efficient and effective GUI Agent, termed \textbf{SimpAgent}.
    To mitigate potential interference from numerous unrelated elements, we introduce a \textbf{masking-based element pruning} method that circumvents the intractable relation modeling through an efficient masking mechanism. 
    To reduce the redundancy in historical information, we devise a \textbf{consistency-guided history compression} module, which enhances implicit LLM-based compression through innovative explicit guidance, achieving an optimal balance between performance and efficiency.
    With the above components, SimpAgent reduces 27\% FLOPs and achieves superior GUI navigation performances. Comprehensive navigation experiments across diverse web and mobile environments demonstrate the effectiveness and potential of our agent.
\end{abstract}    

\vspace{-5mm}
\section{Introduction}

\begin{figure}[t]
    \centering
    \setlength{\abovecaptionskip}{0mm}
    \setlength{\belowcaptionskip}{0mm}
    \includegraphics[width=1.0\linewidth]{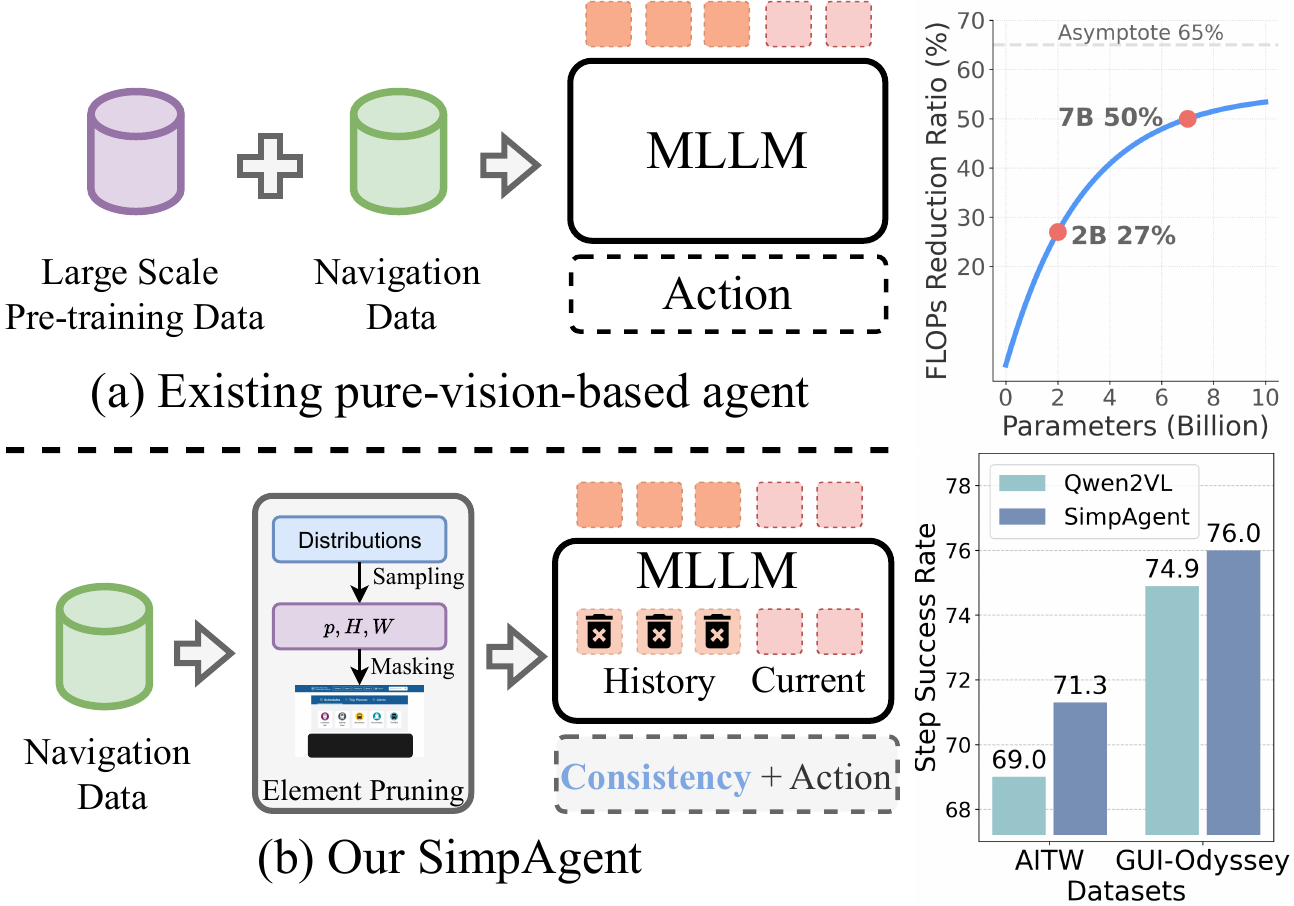}
    \caption{Compared to the large-scale pre-training scheme, we investigate the contextual modeling and propose an efficient training framework with element and history context simplification and additional consistency guidance. The resulting SimpAgent can achieve superior performance with 27\% FLOPs reduction.}
    \label{fig:teaser}
\vspace{-7mm}
\end{figure}

% Building upon large language models (LLMs), GUI agents hold the potential to revolutionize daily task automation, and streamline workflows across various applications \cite{wang2024p,nguyen2024a}. Many current agents \cite{zhang2023h,putta,wen2024,patel2024,chae2024,qi2024,christianos2023,li2025generative} make decisions on textual representations of the environments, such as HTML and accessibility trees, which often require system-level permissions and are often lengthy, noisy, limited scalability \cite{zheng2024d,cheng2024}. An appealing and promising direction is using Multimodal Large language models (MLLMs) to develop GUI agents capable of performing complex tasks simply by analyzing the screenshots \cite{wang2024a,xu2024e,wu2024c,gou2024}.

% These pure-vision-based GUI agents are implemented through either modular agent frameworks \cite{wang2024a,wang2024b,gou2024} or end-to-end agent models \cite{zhang2024f,cheng2024,xu2024e}. The agent framework requires predefined workflows and struggles to handle dynamically changing environments. The end-to-end design presents a unified and integrated architecture, which is more scalable and can easily adapt to new environments \cite{wu2024c}. Due to the significant gap between natural images and GUI screenshots, existing GUI visual agents \cite{gou2024,wu2024c,xu2024e} primarily pursue comprehensive, high-quality training data for improved GUI comprehension ability, which is very valuable but not computation-friendly, as shown in Figure \ref{fig:teaser}. 

Many current GUI agents \cite{zhang2023g,putta,wen2024,patel2024,chae2024,qi2024,christianos2023,li2025generative} make decisions on textual representations of the environments, such as HTML and accessibility trees, which often require system-level permissions and are often lengthy, noisy, limited scalability \cite{zheng2024d,cheng2024}. An appealing and promising direction is using Multimodal Large language models (MLLMs) to develop GUI agents capable of performing complex tasks simply by analyzing the screenshots \cite{wang2024a,xu2024e,wu2024c,gou2024}.
These pure-vision-based GUI agents are implemented through either modular agent frameworks \cite{wang2024a,wang2024b,gou2024,xie2025gui} or end-to-end agent models \cite{zhang2024f,cheng2024,xu2024e}. Due to the significant gap between natural images and GUI screenshots, current pure-vision-based GUI agents primarily pursue comprehensive, high-quality pre-training data for improving GUI comprehension ability, which is very valuable but not computation-friendly, as shown in Figure \ref{fig:teaser}. 

\begin{figure}[t]
    \centering
    \setlength{\abovecaptionskip}{0mm}
    \setlength{\belowcaptionskip}{-2mm}
    \includegraphics[width=1.0\linewidth]{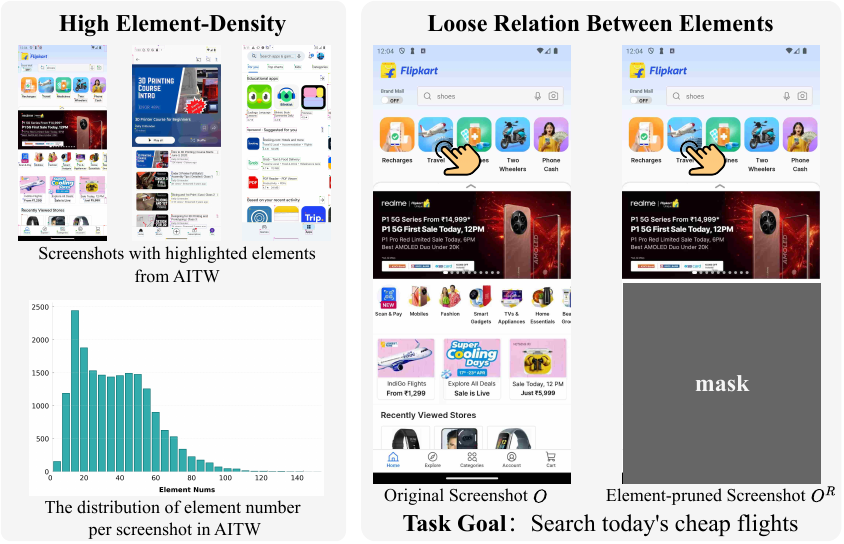}
    \caption{The illustration of high density and loose relation of elements in UI screenshots.}
    \label{fig:pre_ana}
\vspace{-5mm}
\end{figure}

Despite their advancements, building a powerful GUI agent still suffers from significant challenges related to contextual modeling. We analyze the characteristics of element and history contexts to uncover the underlying challenges. \textbf{1) The high-density and loose-relation of element context.} The statistical results in Figure \ref{fig:pre_ana} indicate that the number of elements per screenshot can reach dozens or even hundreds. These elements are not tightly integrated \cite{lowgren2004}, unrelated elements may inadvertently affect each other’s functionality or appearance, hindering the comprehension of critical elements, empirically verified in Table \ref{tab:pre_exp}. \textbf{2) The high redundancy of history context.} Introducing historical observations will increase the computational overhead by 3.4 times, but only improve the performance by 3.0\%, as shown in Table \ref{tab:pre_exp}. The significant gap between computation and performance increments demonstrates a high redundancy in historical information, raising questions about the optimal utilization of historical information.

To address these challenges, we devise a \textbf{context-aware simplification framework} to construct an effective and efficient GUI Agent, termed \textbf{SimpAgent}. The main components within our framework are masking-based element pruning and consistency-guided history compression, enabling the efficient modeling of element and history contexts respectively, depicted in Figure \ref{fig:teaser}. As unrelated elements obstruct the understanding of critical elements, the most effective solution is to eliminate them. By avoiding the challenges of intractable relationship modeling for pruning elements, we successfully reach this goal with an efficient masking mechanism.
% Instead of identifying and pruning them by intractable relationship modeling, we achieve this goal by simply masking.
During the training phase, our \textbf{masking-based element pruning} method randomly masks a rectangular region based on a predefined distribution in the current screenshot. Because unrelated elements occupy a considerable portion of the screenshot, our masking operation can prune them with a high probability. We empirically observe a notable performance improvement even when the masked region reaches half of the screenshot.

To reduce the redundant historical visual information, we devise a \textbf{consistency-guided history compression} method with a token dropping strategy and a consistency guidance. Specifically, we first directly drop historical vision tokens at one specific LLM layer $k$, which will implicitly force LLM to compress historical visual information into preserved tokens. Our dropping strategy shares a similar finding with recent approaches \cite{chen2024n,hu2024d,wen2024b}, the shallow LLM layers can aggregate the vision features into a few anchor tokens (primarily adjacent action tokens in our cases). Compared to existing compression works \cite{li2023m,liu2024k,wen2024b,ye2024a}, our directly dropping strategy requires no extra parameters and enables efficient attention computation. Despite the significant efficiency of dropping, it still encounters the information loss caused by its implicit compression mechanism. Instead of reducing the dropping rate \cite{chen2024n,hu2024d} or adjusting the attention mask \cite{ye2024a}, we innovatively propose a consistency guidance to explicitly steer the compression. We additionally keep an original branch with unchanged tokens after layer $k$, and minimize the Kullback-Leibler divergence between the action predictions from two branches. The experiments indicate that our compression approach achieves a better trade-off between performance and computational efficiency.

% To reduce the redundant historical visual information, we implement a token dropping strategy at one specific LLM layer. Our method is built upon a similar finding with other LLM-based compression approaches \cite{chen2024n,hu2024d, wen2024b}, the shallow LLM layers can aggregate the vision features into a few anchor tokens and the vision tokens are less important in the deep LLM layers. In contrast to existing compression works \cite{li2023m,liu2024k,wen2024b,ye2024a}, our directly dropping method implicitly compresses historical visual information with no extra parameters and enables efficient attention computation. To avoid the information loss from this implicit compression, we add a consistency constraint during the compression process, by maintaining the consistency between the dropping branch and the original branch. The experiments indicate that our consistency-guided history compression method achieves a better trade-off between performance and computational efficiency. We achieve a better balance between performance and efficiency with 27\% FLOPs reduction.

We conduct comprehensive experiments on four representative GUI navigation datasets, AITW \cite{rawles2023}, Mind2Web \cite{deng2023}, GUI-Odyssey \cite{lu2024a}, AndroidControl \cite{li2024a}, to verify the effectiveness and computational efficiency of our methods. The resulting SimpAgent can reduce the inference FLOPs by 27\% and achieves significant performance gains of 2.3\% and 1.1\% on AITW and GUI-Odyssey, respectively. On the AndroidControl dataset, our SimpAgent improves the performance by 0.7\% without any extra pre-training data. In contrast, OS-Altas achieves a performance gain of 0.8\% with 1.9M pre-training GUI data.

Our contributions are summarized as follows:
\begin{enumerate}
    \item We systematically analyze the UI screenshot and GUI Agent from the perspective of contextual modeling, and conclude the characteristics of element and history contexts, highlighting the underlying challenges overlooked in early works.
    \item We propose a context-aware simplification framework with two novel components, masking-based element pruning and consistency-guided history compression, resulting in SimpAgent to efficiently and effectively extract essential elements and historical information.
    \item Comprehensive experiments on four representative GUI navigation datasets, AITW \cite{rawles2023}, Mind2Web \cite{deng2023}, GUI-Odyssey \cite{lu2024a}, AndroidControl \cite{li2024a}, verify the effectiveness and computational efficiency of our SimpAgent. 
\end{enumerate}

\section{Related works}

\textbf{GUI Agents.} 
Autonomous agents powered \cite{yao2022,cheng2024,wang2023h,li2024optimus1,li2025optimus2,li2025star,zhang2024d} by (M)LLMs \cite{liu2023b,li2023,chen2024lion,shen2024mome,li2025lionfs}, have attracted considerable attention owing to their advanced interactive capabilities. 
Recent efforts have begun to enable agents to interact with web, mobile, and desktop operating systems, resulting in GUI agents \cite{wang2024p,nguyen2024a}. However, early works built agents \cite{deng2023,zhang2023g,qi2024,wang2024l} via internal APIs or system backend codes, which is unavailable in most commercial software. 
This has prompted a shift in research focus toward pure-vision-based agents. 
% , which perceive the environmental observation and simulate human-like interactions with digital devices by mimicking mouse and keyboard inputs. 

To leverage the understanding and reasoning capabilities of advanced foundation models (e.g., GPT-4o \cite{hurst2024gpt} and Gemini \cite{team2023gemini}), many works \cite{wang2024a,gou2024,wen2024a} attempt to build compositional, framework-based agents that generally contains multiple models (such as a Planner and an Actor). 
% Uground, Aguvis, and OS-Altas are the representative works in this research line. 
They collect or synthesize a large-scale GUI grounding corpus to develop strong foundation action models and combine them with an advanced Planner (based on GPT-4o or Gemini). Some works \cite{xu2024e,qin2025} also curate lots of GUI navigation datasets to further build end-to-end native agents. Despite their advancements, the collection of such massive datasets requires substantial human and computational resources, which can have huge economic impacts.

Instead of massive data curation, a few works have attempted to construct effective and computation-friendly agents. OdysseyAgent \cite{lu2024a} proposes a history resampler module to compress historical screenshots, but it overlooks the multi-modal interaction and introduces extra parameters compared to our method. Iris \cite{ge2024} and ShowUI \cite{lin2024d} try to eliminate the useless background by leveraging low-level visual clues in UI screenshots, but they struggle to handle element-density screenshots. In contrast, we devise an efficient masking-based element-pruning method to enhance the comprehension of critical elements even in high element-density scenarios.

\noindent\textbf{Vision Compression in MLLM.} 
In the real world, multimodal content understanding plays a crucial role and holds significant value \cite{shtedritski2023,shao2023detecting,shao2024detecting,shao2019multi}. This capability has been further enhanced and emphasized in multimodal large language models.
As the number or resolution of images increases, the vision tokens in MLLMs will take up a significant portion of the limited context window and incur huge computational costs. To address it, previous works \cite{li2023,li2023m,hu2023,dai2023,zhang2025falcon} mainly rely on a set of learnable queries to compress vision tokens, which requires a multi-stage dedicated training procedure. Some works \cite{shang2024,liu2024k,li2024c} try to use token similarity to select vision tokens, but still introduce extra learnable parameters. Recently, FastV \cite{chen2024n} has found that shallow LLM layers can aggregate visual features to some anchor tokens.
% , resulting in an attention-score-based token pruning method for inference acceleration with inevitable performance degradation. 
VoCo-LLaMA \cite{ye2024a} adjusts the attention masking to explicitly compress visual information into VoCo tokens, preventing efficient attention implementation (like FlashAttention).
% To improve training efficiency, 
Victor \cite{wen2024b} keeps the attention masking unchanged and implicitly summarizes visual information into some registers. In contrast to them, we propose direct token dropping and consistency guidance to achieve a better trade-off between performance and efficiency.

\vspace{-3mm}
\section{Preliminary}
% \quad In this section, we will first formulate the GUI Agent and GUI Navigation tasks, then present some analyses about the characteristics of element and history context in GUI Agent and underscore the key challenges.

\subsection{Problem Definition}
GUI Agents, which are constructed for GUI Navigation tasks, can sequentially generate actions and interact with the GUI environment to achieve a task goal \cite{nguyen2024a}. The interaction between an agent and the environment can be treated as a sequential decision-making process \cite{wang2024p}. At time step $t$, the agent receives a task goal $G$, a history context $H_{t}$, a screenshot observation $o_t$, and then generates an action $a_t$. The history context contains previous screenshot observations and actions, $H_{t} = \{o_{t-\tau},a_{t-\tau},\dots, o_{t-1},a_{t-1}\}$, where $\tau$ is history length, providing valuable multimodal information about the task completion status. If we represent the agent model as $\pi_{\theta}$, the training objective of the agent can be formulated as follows:

\begin{equation}
    \mathcal{L}(\pi_{\theta}) = - \sum_{t=0}^{T} \log \pi_{\theta}(a_t | o_t, H_t, G).
\end{equation}

% The action space is a finite operation command set with dynamic parameters. The screen observation space contains all possible designs from various applications.

\subsection{The high-density and loose-relation of element context}
Perceiving the GUI screenshots is critical and challenging for the agent's decision-making, due to the high element-density, intricate layout, and diverse styles of the interface \cite{cheng2024,wu2024c}. Figure \ref{fig:pre_ana} shows some screenshots on the mobile device, providing an intuitive representation of high-element density in the GUI scenario. Statistical analyses reveal that each screenshot in the AITW dataset \cite{rawles2023} contains an average of 56 elements, while the AndroidControl dataset \cite{li2024a} has an average of 180 elements per screenshot. 

Another distinctive characteristic of graphical elements is their loose relations. The associations between elements are often loose and changing, replacing, or hiding one element may not significantly impact the user's ability to interact with or understand other elements. The main reason for these loose relations is the modular interface design principle, where components are designed to function independently \cite{lowgren2004}. In Figure \ref{fig:pre_ana}, we present a case that masking bottom regions (including unrelated elements) in a screenshot has no impact on the decision-making compared to the usage of the original screenshot. 

\begin{table}[t]\small
    \centering
    \setlength{\abovecaptionskip}{0mm}
    \setlength{\belowcaptionskip}{0mm}
    \begin{tabular}{lcc|c}
    \toprule[1.2pt]
      Settings  & \#Tokens & FLOPs(T) (All / LLM) & Step SR \\
    \midrule
      $O$    & 327 & 2.45 / 0.88 &  62.5  \\
      $O + 4A$    & 408 & 2.70 / 1.09 & 66.0 \\
      $O^R + 4A$ & 408 & 2.70 / 1.09 & 68.8 \\
      $O + 4AO$   & 1551 & 11.90 / 4.12 & 69.0 \\
    \bottomrule[1.2pt]
    \end{tabular}
    \caption{The pilot experiments of GUI Agent on the AITW dataset. The base model is Qwen2-VL-2B. ``$O$" (``$O^R$") means only using current observation (pruned unrelated elements) to make a decision, while ``$+4AO$" or ``$+4A$" denotes additional adding 4 previous observation-actions or actions.}
    \label{tab:pre_exp}
\vspace{-5mm}
\end{table}

To further validate the loose relation between elements, we conduct a preliminary experiment by roughly splitting a screenshot observation $O$ into two sets, the areas with related elements $O^R$ and unrelated elements $O^U$. As each observation corresponds to an action (mostly a click operation), we assume that the areas far from the click point contain unrelated elements. Therefore, we consider half of the image containing the click point as $O^R$, and the rest is $O^U$. Table \ref{tab:pre_exp} show the decision-making performance (68.8\%) on $O^R$ is superior to that (66.0\%) on $O$. The comparison confirms the existence of unrelated elements and highlights the potential interference they may cause.

\begin{figure*}[h]
    \centering
    \setlength{\abovecaptionskip}{0mm}
    \setlength{\belowcaptionskip}{0mm}
    \includegraphics[width=1.0\textwidth]{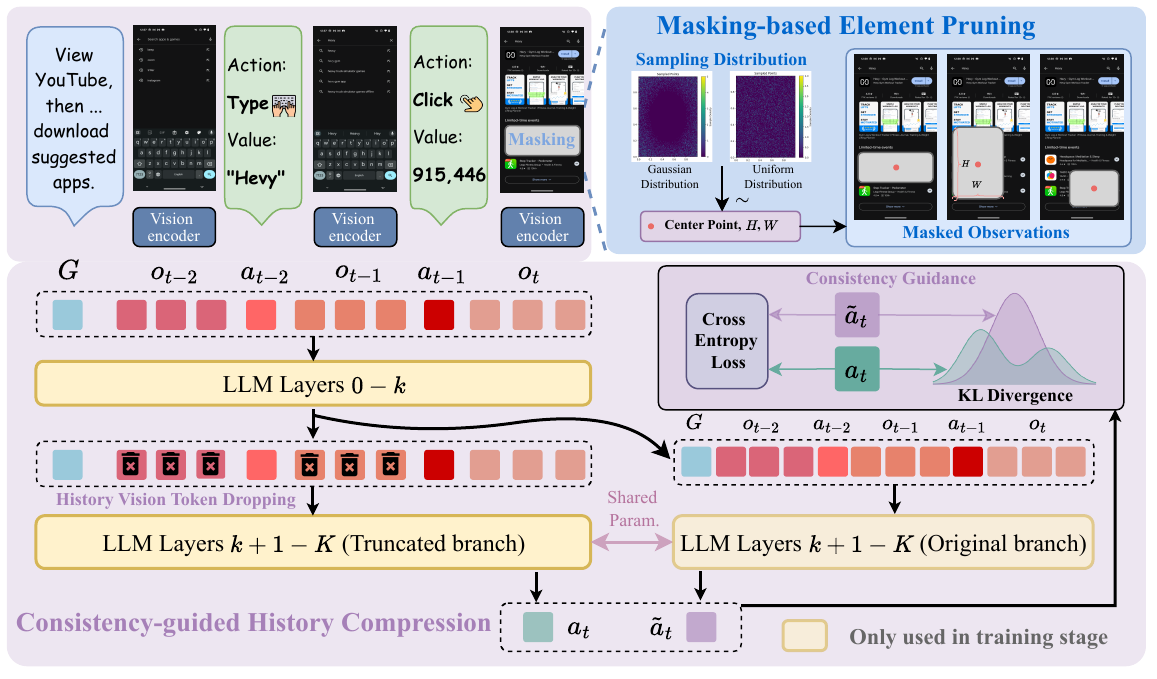}
    \caption{Overview of our context-aware simplification framework for building SimpAgent. The main components are masking-based element pruning and consistency-guided history compression.
    % These two methods simplify a GUI agent model from two aspects, element and history contexts, respectively.
    We assist SimpAgent in mitigating interference from unrelated elements by masking certain regions based on a pre-defined distribution.
    During training, we maintain the consistency between two LLM branches for explicitly steering history compression.
    % , incurring 10\% extra computational costs.
    At inference, SimpAgent only uses the LLM branch with truncated tokens, reducing 27\% FLOPs.}
    \label{fig:framework}
\vspace{-5mm}
\end{figure*}

\subsection{The high redundancy of history context}
The integration of historical information, including previous observations and actions, can help the agent to comprehend the completion status of the current task. As many GUI navigation tasks are quite complex and require a lot of steps to complete, effectively using history context is very valuable \cite{zhang2024f,lu2024a}. Previous works \cite{zhang2024f,ma2024d} mainly leverage the prior actions, which leads to considerable gains and a few extra tokens (generally 10-20 tokens per action). However, the vision information from previous observations is also important because the task progress can be accurately determined only with the combination of previous observations and actions \cite{lin2024d}.

Here, we present some pilot experiments of history observations and actions by fine-tuning Qwen2-VL-2B \cite{wang2024n} as an agent on the popular GUI navigation task, AITW \cite{rawles2023}. In Table \ref{tab:pre_exp}, we compare the agent's performance with different levels of history context, without history, with 4 previous actions, and with 4 previous actions and observations. It is clear that current observation has a dominant impact on the final outcome, only using current observation achieves 62.5\% step success rate. Incorporating 4 actions can increase the step success rate by $3.5\%$, and further adding 4 observations leads to the gain of $3.0\%$. The current observation setting only requires 327 tokens, while the 4 observations additionally introduce 1143 tokens and increase the computational overhead by 3.4 times. The comparison of performance and computational increments indicates a significant redundancy in historical observation information, which can notably impact the agent's inference speed and hinder its practical application.
% 强调 history vision 和 current vision 之间的差异，来说明 history redundancy.

\section{A context-aware simplification framework}
% \vspace{-3mm}
In this section, we will introduce our context-aware simplification framework with two novel methods, masking-based element pruning and consistency-guided history compression, as shown in Figure \ref{fig:framework}.
% , for building an effective and efficient GUI agent, namely SimpAgent.
% Our simplification framework assists the agent in extracting essential information while eliminating redundant or disruptive elements that could hinder the agent's understanding.

\subsection{Masking-based Element Pruning}
As analyzed above, the characteristic of high element-density in UI screenshots results in a significant challenge in perceiving valuable elements.
% Current works \cite{gou2024,wu2024c} mainly pursue the scaling of grounding pre-training data to enhance the comprehension ability of complex and diverse elements.
% They manually or automatically collect large-scale screenshots and associate elements in screenshots with a referring expression for grounding pre-training. 
To address it, current works \cite{gou2024,wu2024c} mainly pursue the scaling of grounding pre-training data, which is effective but very expensive due to the cumbersome data collection process. In contrast, we adopt an orthogonal approach to elevate element comprehension capability in a data-efficient manner by leveraging the inherent characteristics of user interfaces (UIs).
% Despite the inherent challenges posed by high element-density, the existence of loose relations between elements provides a foundation for pursuing a viable and effective solution strategy. 
Our method is inspired by the implication of loose relations, that pruning some unrelated elements plausibly has no impact on the comprehension of key elements. However, identifying what elements to prune is intractable cause the relation between elements is dependent on numerous factors, highly unpredictable \cite{lowgren2004}. To avoid the modeling of element relation, we devise a conceptually simple but effective method to prune elements by masking.

Our masking-based element pruning method selects a rectangle region $\mathcal{R}$ in current observation $o_t$, and performs a masking operation $\mathcal{M}$ on it. Assume the size of the selected region is $h\times w$, we randomly sample the size from a uniform distribution $U(a,b)$ where $a$, $b$ are left and right bounds. The center point of the selected region is $p_c=(x_c,y_c)$, we randomly sample this point from a specific distribution (Gaussian or Uniform distribution). The masking operation will set the pixels within $\mathcal{R}$ to a specific value $v$. The masking method and examples are represented in Figure \ref{fig:framework}.  In training, our masking operation is conducted with a certain probability $p$. At inference, the masking operation is omitted to provide complete visual information.

\begin{equation}
    o^m_t = \mathcal{M}(o_t) = 
    \begin{cases}
    v, & (x,y) \in \mathcal{R}; \\
    o_t(x,y), & otherwise.
    \end{cases}
\end{equation}

Our core design principle is that the masking of unrelated elements can facilitate the comprehension of valuable elements. Our method can prune unrelated elements with a high probability for two reasons. 1) The most valuable elements cover a small area in the screenshot. we have conducted a statistical analysis and found that the average portion of click regions in the screenshot is $2\%$. 2) The loose relation between elements implies that unrelated elements occupy a considerable portion of the screenshot. These two reasons collectively ensure that the masking operation will not yield a large quantity of noise samples. In the experiment section, we find that notable performance improvements can still be achieved even when the masked region reaches half of the screenshot.

\subsection{Consistency-guided History Compression}

The high redundancy of history context requires us to compress the historical information and improve the inference efficiency of the agent. In comparison with few computation cost caused by the history actions, history observations additionally increase the computational overhead by 3.4 times. Thus, we focus on the compression of history visual information. Current vision compression approaches \cite{li2023m,liu2024k,ye2023a} mainly employ the extra-module-based paradigm, which will induce additional computational cost and hardly take into account the interaction between multiple modalities. Recently, some researchers \cite{chen2024n,hu2024d,wen2024b} investigate the attention distribution of various types of tokens in a trained MLLM. They found that the shallow layers in LLM can aggregate the vision features into a few anchor tokens and the attention scores can be used to identify these anchor tokens. These works can only retain the anchor tokens to accelerate the inference of a trained MLLM, but inevitably encounter performance degradation. In contrast, we propose a consistency-guided history compression method and apply it in the training stage to achieve a better trade-off between performance and computational efficiency.

% Therefore, a few works try to explore LLM-based token compression during the training stage to achieve a better trade-off between performance and efficiency. Inspired by them, we propose consistency-guided history compression methods to reduce the computational overhead resulting from the history observations while maintaining nearly lossless performance.

\begin{figure}
    \centering
    \setlength{\abovecaptionskip}{0mm}
    \setlength{\belowcaptionskip}{0mm}
    \includegraphics[width=1.0\linewidth]{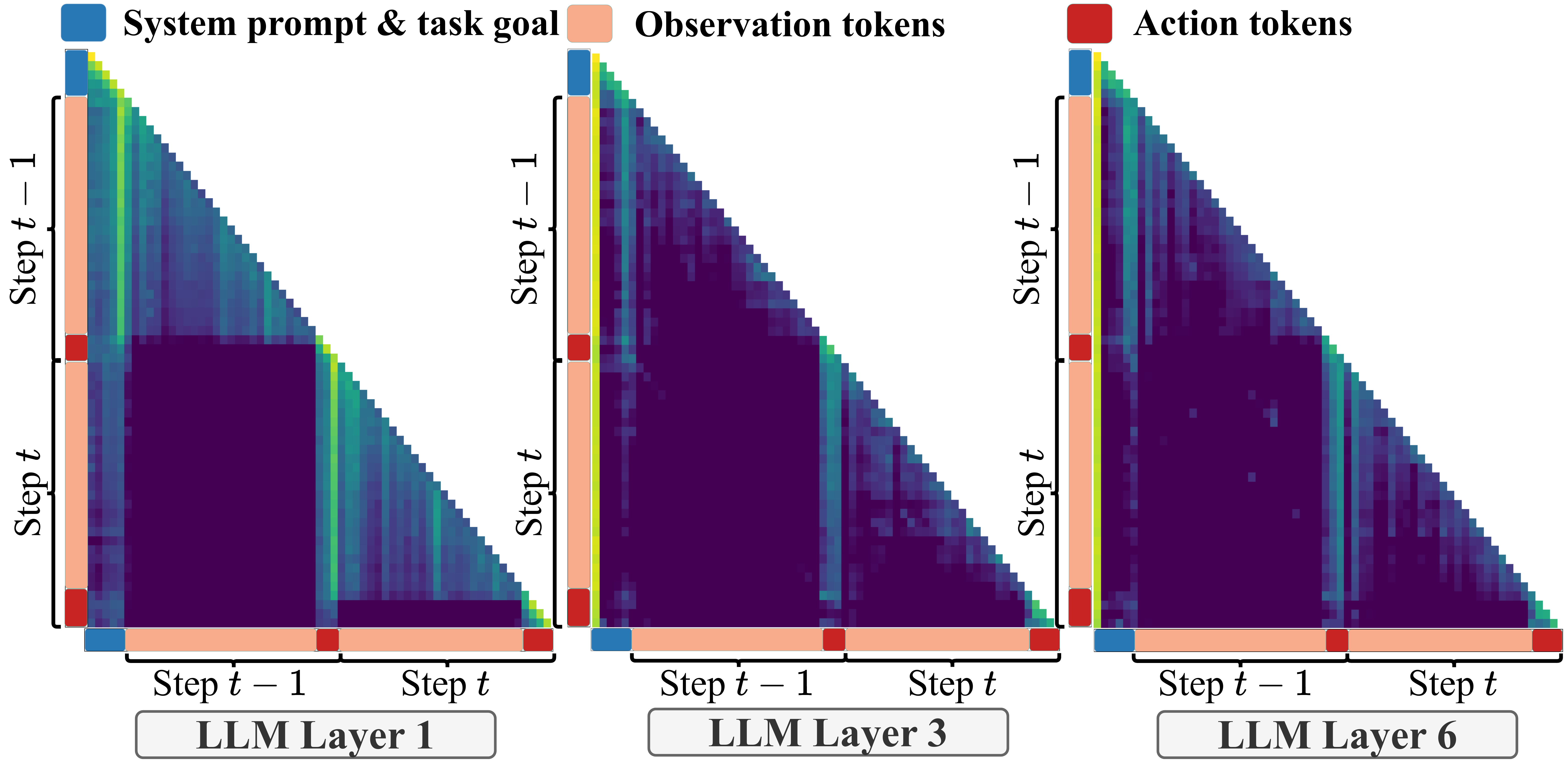}
    \caption{The attention maps during the LLM decoding process of GUI Agent. One previous step is used. We can see that in the shallow layers, attention distributes smoothly across various types of tokens within each step. In the deep layers, the attention scores within each step are aggregated to adjacent action tokens and attention over observation tokens is rather sparse.}
    \label{fig:ana_att_method}
\vspace{-5mm}
\end{figure}

% introduce our method, 我们历史压缩方法
Our consistency-guided history compression contains two key components, LLM-based history dropping and consistency-guided training objective. The history compression is internally processed within the LLM through the selective removal of vision tokens. After projecting task goal, raw observations, and actions to vision and text tokens, the concatenated token sequence $\{G,\underbrace{o_{t-\tau}, a_{t-\tau}, \dots, o_{t-1}, a_{t-1}}_{H_t},o_t\}$ is fed into a LLM to generate next action tokens $a_t$. We simply drop all history vision tokens $\{o_{t-\tau},\dots,o_{t-1}\}$ after LLM layer $k$, and the LLM continues with the truncated token sequence $\{G,\underbrace{a_{t-\tau},\dots,a_{t-1}}_{H^c_t},o_t\}$. The whole process is depicted in Figure \ref{fig:framework}. For simplicity, we represent the agent model with LLM-based history dropping as $\pi_{\theta}(a_t | o^m_t, H^c_t, G)$. The rationale behind this approach is that shallow layers in LLM can use causal self-attention to compress vision tokens into adjacent text tokens, which is also observed in Figure~\ref{fig:ana_att_method}.

We find that the directly dropping strategy implicitly compresses the visual information through the causal self-attention mechanism, inevitably leading to some information loss. VoCo-LLaMA \cite{ye2024a} introduces an explicit and strict compression method by adjusting the attention mask but suffers from severe computational efficiency issues. In contrast, we devise a consistency-guided training objective to steer the vision compression process by maintaining the consistency between the truncated branch (with $H^c_t$) and the original branch (with $H_t$). The training objective is formulated as follows:
% \begin{equation}
% \begin{aligned}
% % \begin{multline}
%     \mathcal{L}(\pi_{\theta}) &= - \mathbb{D}_{KL} [\pi_{\theta}(\tilde{a}_t | o^m_t, H_t, G) \| \pi_{\theta}(a_t | o^m_t, H^c_t, G)]  \\
%     &\quad - \sum_{t=0}^{T} \log \pi_{\theta}(\tilde{a}_t | o^m_t, H_t, G) \\
%     &\quad - \sum_{t=0}^{T} \log \pi_{\theta}(a_t | o^m_t, H^c_t, G),
% \end{aligned}
% % \end{multline}
% \end{equation}
% \vspace{-3mm}
\begin{equation}
\begin{aligned}
% \begin{multline}
    \mathcal{L}(\pi_{\theta}) = - \mathbb{D}_{KL} [\pi_{\theta}(\tilde{a}_t | o^m_t, H_t, G) \| \pi_{\theta}(a_t | o^m_t, H^c_t, G)]  \\
    \quad - \sum_{t=0}^{T} \log \pi_{\theta}(\tilde{a}_t | o^m_t, H_t, G)
    - \sum_{t=0}^{T} \log \pi_{\theta}(a_t | o^m_t, H^c_t, G).
\end{aligned}
% \end{multline}
\end{equation}
\vspace{-3mm}

% discuss the difference from the most related works, and its advantages and key properties.
% idea flop reduction ratio!
% Compared to current state-of-the-art LLM-based compression methods [c], our approach does not require token selection using the attention matrix, enabling efficient attention implementation like FlashAttention. While Victor [c] shares the same advantage, it introduces additional summarization tokens that require more parameters and incur higher computational costs. 
% [highlighting computational cost analysis.]

\section{Experiments}

In this section, we validate our model using comprehensive agent benchmarks that cover various mobile scenarios. We use Qwen2VL-2B \cite{wang2024n} as our base model and fine-tune it on GUI Navigation datasets to obtain our SimpAgent. The implementation details can be found in section B in Appendix.
% The results on different datasets demonstrate the effectiveness of our approach. To showcase the superiority of our compression method, we compared it with existing compression techniques on AITW[cite] dataset, achieving optimal performance and efficiency. Subsequently, in the ablation studies, we analyzed the effectiveness of each module design and the stability under different configurations.

\subsection{Datasets}
We select four GUI navigation datasets, to evaluate SimpAgent's multi-step execution ability. Table \ref{tab:dataset_info} shows the statistical analysis of these datasets.

% \textbf{Android In The Wild} (AITW) \cite{rawles2023} consists of 30k instructions and 715k operation trajectories in the context of smartphone environments. However, the original train-test split carries the risk of overfitting due to overlapping instructions between the training and test sets, as well as the presence of multiple similar trajectories per instruction \cite{cheng2024}. Therefore, we adopt the instruction-wise split scheme proposed by SeeClick \cite{cheng2024} for the AITW dataset to achieve a fair evaluation.

% \textbf{Mind2Web} \cite{deng2023} comprises over 2000 open-ended tasks collected from 137 real websites, originally designed for text-based agents. We use the processed data and follow the same setting in SeeClick.

% \textbf{GUI-Odyssey} \cite{lu2024a} is a comprehensive dataset for training and evaluating cross-app navigation agents, consisting of 7,735 episodes from 6 mobile devices. We follow the same evaluation setting in the original paper.

% \textbf{AndroidControl} \cite{li2024a} encompasses 14,548 unique tasks across 833 Android apps, offering both high and low-level instructions to explore the level of task complexity an agent can handle. We follow the same evaluation setting in OS-Atlas \cite{wu2024c}.

% \subsection{Implementation Details}
% \quad We use Qwen2VL-2B as our base model. For different datasets, we adopt the same data organization format following SeeClick[site]. We parse the executed actions in JSON format, dividing them into two parts: action\_type and action\_value.

\begin{table}[h]\footnotesize
\centering
\setlength{\abovecaptionskip}{0mm}
\begin{tabular}{lccc}
\toprule[1.2pt]
Dataset & Domain & \#Task & \#Avg. Steps \\
\hline
AITW \cite{rawles2023} & Mobile \& Web & 2,939 & 8.1 \\
Mind2Web \cite{deng2023} & Web & 2,350 & 7.3 \\
GUI-Odyssey \cite{lu2024a} & Mobile & 7,735 & 15.4 \\
AndroidControl \cite{li2024a} & Mobile & 15,283 & 5.5 \\
\bottomrule[1.2pt]
\end{tabular}
\caption{Dataset statistics, including domain, number of tasks, and average task length (measured in steps).}
\label{tab:dataset_info}
\vspace{-5mm}
\end{table}

\subsection{Main Results}
In Tables \ref{tb:main_amg}, we present the evaluation results of our SimpAgent on mobile and web navigation datasets, AITW, Mind2Web, GUI-Odyssey. In comparison with SeeClick \cite{cheng2024}, Iris \cite{ge2024}, and ShowUI \cite{lin2024d}, our agent demonstrated superior performance without pre-training GUI datasets and fewer or comparable parameters. We also evaluate our SimpAgent on a long-horizon navigation dataset, GUI-Odyssey with average steps of 15.4. In the last column of Table \ref{tb:main_amg}, we find that SimpAgent outperforms OdysseyAgent \cite{lu2024a} with fewer parameters, suggesting its superior capability of history information utilization. In Table \ref{tb:main_ac}, we evaluate our model on the most diverse mobile navigation dataset, AndroidControl, with 833 Android apps. We adopt the setting of only using high-level instructions. Notably, with no extra pre-training data and smaller model size, SimpAgent-2B yields comparable performance gain (increase 0.7\% over the baseline Qwen2VL) with OS-Atlas-4B \cite{wu2024c} (increase 0.8\% over the baseline InternVL2) which has been pre-trained on large-scale GUI datasets (1.9M). These comparisons suggest that our context-aware simplification framework can assist agents in effectively extracting critical elements and enhancing the agent's navigation capability. We also evaluate our SimpAgent and its variants on the Mind2Web dataset with three challenging out-of-the-domain testing sets, obtaining new state-of-the-art performances. \textbf{More detailed comparisons can be found in Table 1-3 in Appendix.}

\begin{table*}[h]\small
\centering
\setlength{\abovecaptionskip}{0mm}
% \resizebox{1.0\textwidth}{!}{%\renewcommand\arraystretch{1.1}
\begin{tabular}{lccccccc}
\toprule[1.2pt]
\multirow{2}{*}{Method} & \multirow{2}{*}{\#P.T.} & \multirow{2}{*}{\#Param.} & \multirow{2}{*}{AITW}  & \multicolumn{3}{c}{Mind2Web}  & \multirow{2}{*}{GUI-Odyssey} \\ 
\cmidrule(lr){5-7}
  & & & & Cross-Task & Cross-Website & Cross-Domain & \\
\midrule
CogAgent \cite{hong2023} & 140M & 18B & - & 17.6 & 13.4 & 15.5 & 11.84 \\
Qwen-VL \cite{bai2023} & - & 9.6B & 54.3 & 13.3 & 9.2 & 12.0 & 72.8 \\
SeeClick \cite{cheng2024} & 850K & 9.6B & 59.3 & 25.5 & 16.4 & 20.8 & - \\
OdysseyAgent \cite{lu2024a} & - & 9.6B & - & - & - & - & 74.3 \\
Qwen2VL$^{*}$ \cite{wang2024n} & - & 2B & 66.0 & 45.0 & 40.9 & 40.5 & 69.0 \\
Qwen2VL \cite{wang2024n} & - & 2B & 69.0 & 46.7 & 42.2 & 44.6 & 74.9 \\
Iris \cite{ge2024} & 850K & 9.6B & 63.6 & 32.0 & 26.2 & 28.8 & - \\
ShowUI \cite{lin2024d} & 256K & 2B & 70.0 & 37.2 & 35.1 & 35.2 & - \\
\rowcolor[HTML]{E7EEFE}
SimpAgent & - & 2B & 71.3 & 47.1 & \textbf{42.8} & 43.3 & 76.0 \\ 
\rowcolor[HTML]{E7EEFE}
SimpAgent-M & - & 2B & \textbf{71.5} & \textbf{48.7} & 42.2 & \textbf{45.0} & \textbf{77.4} \\ 
\bottomrule[1.2pt]
\end{tabular}
% }
\caption{Performance comparisons on AITW, Mind2Web, and GUI-Odyssey. We report the step success rate (step SR). ``\#P.T.", and ``\#Param." denote the number of pre-training GUI datasets and parameters, respectively. ``*" means the variant utilizing only action history. ``-M" denotes only applying masking-based element pruning without inference FLOPs reduction.}
\label{tb:main_amg}
\vspace{-5mm}
\end{table*}

% \begin{table*}[t]
% \centering
% \caption{Main Results on Android Control. For Android Control, we report two settings (Low and High).}
% \label{tb:main_ac}
% % \resizebox{1.0\textwidth}{!}{%\renewcommand\arraystretch{1.1}
% \begin{tabular}{lcccccc}
% \toprule[1.2pt]
% \multirow{2}{*}{Method}    & \multicolumn{3}{c}{AndroidControl-Low}  & \multicolumn{3}{c}{AndroidControl-High} \\ 
% \cmidrule(lr){2-4} \cmidrule(lr){5-7}
%  & Type & Grounding & SR & Type & Grounding & SR \\
% \hline
% SeeClick & 93.0 & 73.4 & 75.0 & 82.9 & 62.9 & 59.1 \\
% Qwen2-VL-2B &  &  &  &  &  &  \\
% InternVL-2-4B & 90.9 & 84.1 & 80.1 & 84.1 & 72.7 & 66.7 \\
% OS-Atlas-4B & 91.9 & 83.8 & 80.6 & 84.7 & 73.8 & 67.5 \\
% Qwen2-VL-7B & 91.9 & 86.5 & 82.6 & 83.8 & 77.7 & 69.7 \\
% OS-Atlas-7B & 93.6 & 88.0 & 85.2 & 85.2 & 78.5 & 71.2 \\
% \rowcolor[HTML]{E7EEFE}
% SimpAgent-2B  &  &  &  & 84.4 & 72.1 & 67.9 \\
% \bottomrule[1.2pt]
% \end{tabular}
% % }
% \end{table*}

\begin{table}[h]\footnotesize
\centering
\setlength{\abovecaptionskip}{0mm}
% \resizebox{1.0\textwidth}{!}{%\renewcommand\arraystretch{1.1}
\begin{tabular}{lccccc}
\toprule[1.2pt]
\multirow{2}{*}{Method} & \multirow{2}{*}{\#P.T.} & \multirow{2}{*}{\#Param.} & \multicolumn{3}{c}{AndroidControl-High} \\ 
\cmidrule(lr){4-6}
 & & & Type & Grounding & SR \\
\hline
SeeClick \cite{cheng2024} & 850K & 9.6B & 82.9 & 62.9 & 59.1 \\
InternVL2 \cite{chen2023k} & - & 4B & 84.1 & 72.7 & 66.7 \\
OS-Atlas \cite{wu2024c} & 1.9M & 4B & 84.7 & \textbf{73.8} & 67.5 \\
% Qwen2VL & - & 7B & 83.8 & 77.7 & 69.7 \\
% OS-Atlas & 1.9M & 7B & 85.2 & 78.5 & 71.2 \\
Qwen2VL$^{*}$ \cite{wang2024n} & - & 2B & 84.4 & 72.2 & 67.8 \\
Qwen2VL \cite{wang2024n} & - & 2B & 84.5 & 72.9 & 68.4 \\
\rowcolor[HTML]{E7EEFE}
SimpAgent & - & 2B & \textbf{84.9} & 73.2 & 69.1 \\
\rowcolor[HTML]{E7EEFE}
SimpAgent-M & - & 2B & 84.8 & \textbf{73.8} & \textbf{69.3} \\
\bottomrule[1.2pt]
\end{tabular}
% }
\caption{Main Results on Android Control. ``\#P.T.", ``\#Param." and ``SR" denote the number of pre-training GUI datasets and parameters, and step success rate, respectively. ``*" means the variant utilizing only action history. ``-M" denotes only applying masking-based element pruning without inference FLOPs reduction.}
\label{tb:main_ac}
\vspace{-5mm}
\end{table}

\subsection{Ablation Studies}

\begin{table}[h]\footnotesize
    \centering
    \setlength{\abovecaptionskip}{0mm}
    \begin{tabular}{ccc|l|ll}
    \toprule[1.2pt]
        C.p. & C. C.p. & M. P. & FLOPs & AITW  & GUI-Odyssey\\
    \midrule
        % 4A &  &  &  & 2.70 & 66.0 & 69.0\\
          &  &  & 11.90 & 69.0 & 74.9\\
         \Checkmark &  &  & 8.71$_{\textcolor{teal}{\downarrow\textbf{\scriptsize27\%}}}$ & 67.3$_{\textcolor{maroon}{\downarrow\textbf{\scriptsize1.7}}}$ & 71.8$_{\textcolor{maroon}{\downarrow\textbf{\scriptsize3.1}}}$\\
         
         \Checkmark & \Checkmark &  & 8.71$_{\textcolor{teal}{\downarrow\textbf{\scriptsize0\%}}}$ & 68.9$_{\textcolor{teal}{\uparrow\textbf{\scriptsize1.6}}}$ & 73.7$_{\textcolor{teal}{\uparrow\textbf{\scriptsize1.9}}}$\\
         
         \Checkmark & \Checkmark & \Checkmark & 8.71$_{\textcolor{teal}{\downarrow\textbf{\scriptsize0\%}}}$ & \textbf{71.3}$_{\textcolor{teal}{\uparrow\textbf{\scriptsize2.4}}}$ & \textbf{76.0}$_{\textcolor{teal}{\uparrow\textbf{\scriptsize2.3}}}$\\
    \bottomrule[1.2pt]
    \end{tabular}
    \caption{Ablation study of all components in our context-aware simplification framework. ``C.p." and ``C. C.p." mean history compression and consistency-guided history compression. ``M. P." denotes masking-based element pruning.}
    \label{tab:ablation_study}
\vspace{-5mm}
\end{table}

\textbf{Ablation of all components.} 
% There are two types of historical information utilized: previous actions and historical screenshots. 4A refers to using only the 4 most recent action texts, while 4AO includes both the 4 most recent action texts and historical screenshots. 
In Table \ref{tab:ablation_study}, we examine all components in our context-aware simplification framework. The comparisons begin with history compression, followed by the addition of consistency guidance, and conclude with masking-based element pruning. As shown, if historical screenshots are compressed by directly dropping, the information loss is significant (decreases 1.7\% and 3.1\% on AITW and GUI-Odyssey respectively) due to the implicit supervision mechanism. However, the compression results in a significant inference acceleration and achieves 27\% FLOPs reduction. By utilizing consistency loss to explicitly steer the process with complete historical information, the compression of historical visual information can be significantly enhanced, achieving nearly lossless results (a gap of 0.1\%) on AITW and competitive performance (a gap of 1.2\%) on GUI-Odyssey. The masking-based element pruning method further enhances performance by filtering out irrelevant element noise during the learning process, leading to superior results (increase 2.4\% and 2.3\% on AITW and GUI-Odyssey separately). This demonstrates the effectiveness of our proposed masking-based element pruning and consistency-guided history compression methods.

% \begin{table}[h]\small
%     \centering
%     \setlength{\abovecaptionskip}{1mm}
%     \begin{tabular}{c|cccc}
%     \toprule[1.2pt]
%        $(a,b)$ & $(0.1,0.3)$ & $(0.3,0.5)$ & $(0.5,0.7)$ & $(0,0)$ \\
%     \midrule
%        Step SR  & 70.7 & 70.6 & \textbf{71.3} & 69.0 \\
%     \bottomrule[1.2pt]
%     \end{tabular}
%     \caption{The analysis of masking region ($h,w \sim U(a,b)$) in masking-based element pruning module. We fix the masking probability to $0.5$.}
%     \label{tab:mask_ana}
%     \vspace{-3mm}
% \end{table}

% \begin{figure}[h]
%     \centering
%     \setlength{\abovecaptionskip}{0mm}
%     \includegraphics[width=0.8\linewidth]{figs/ana_masking.png}
%     \caption{The analysis of masking region ($h,w \sim U(a,b)$) in masking-based element pruning module.}
%     \label{fig:mask_ana}
%     \vspace{-3mm}
% \end{figure}

\begin{figure}[h]
    \centering
    \setlength{\abovecaptionskip}{0mm}
    \begin{minipage}[t]{0.48\linewidth}
    \vspace{0pt}
        \centering
        \includegraphics[width=\linewidth]{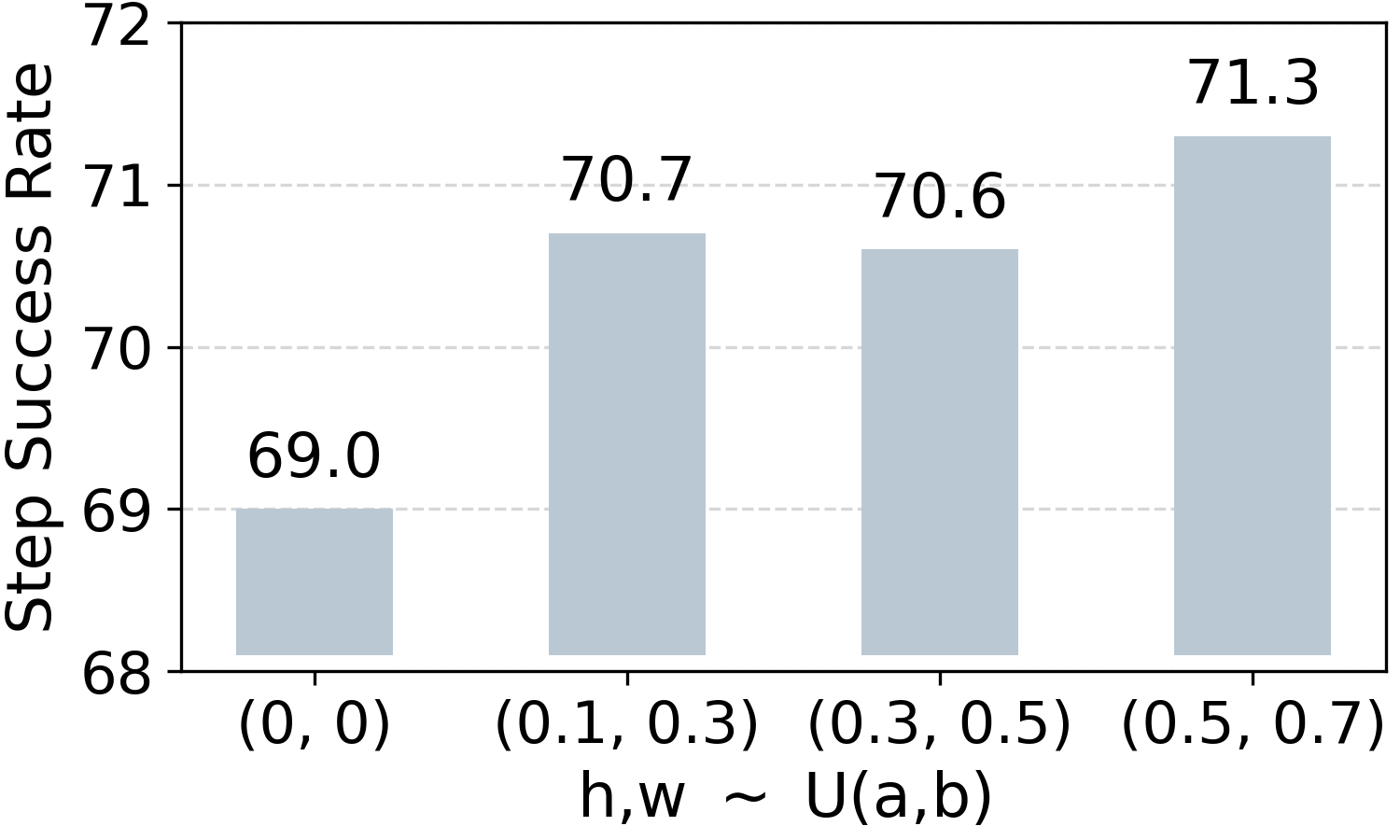}
        \caption*{(a)}
    % \label{fig:mask_ana}
    \end{minipage}
    \hfill
    \begin{minipage}[t]{0.48\linewidth}
    \vspace{0pt}
        \centering
        \includegraphics[width=\linewidth]{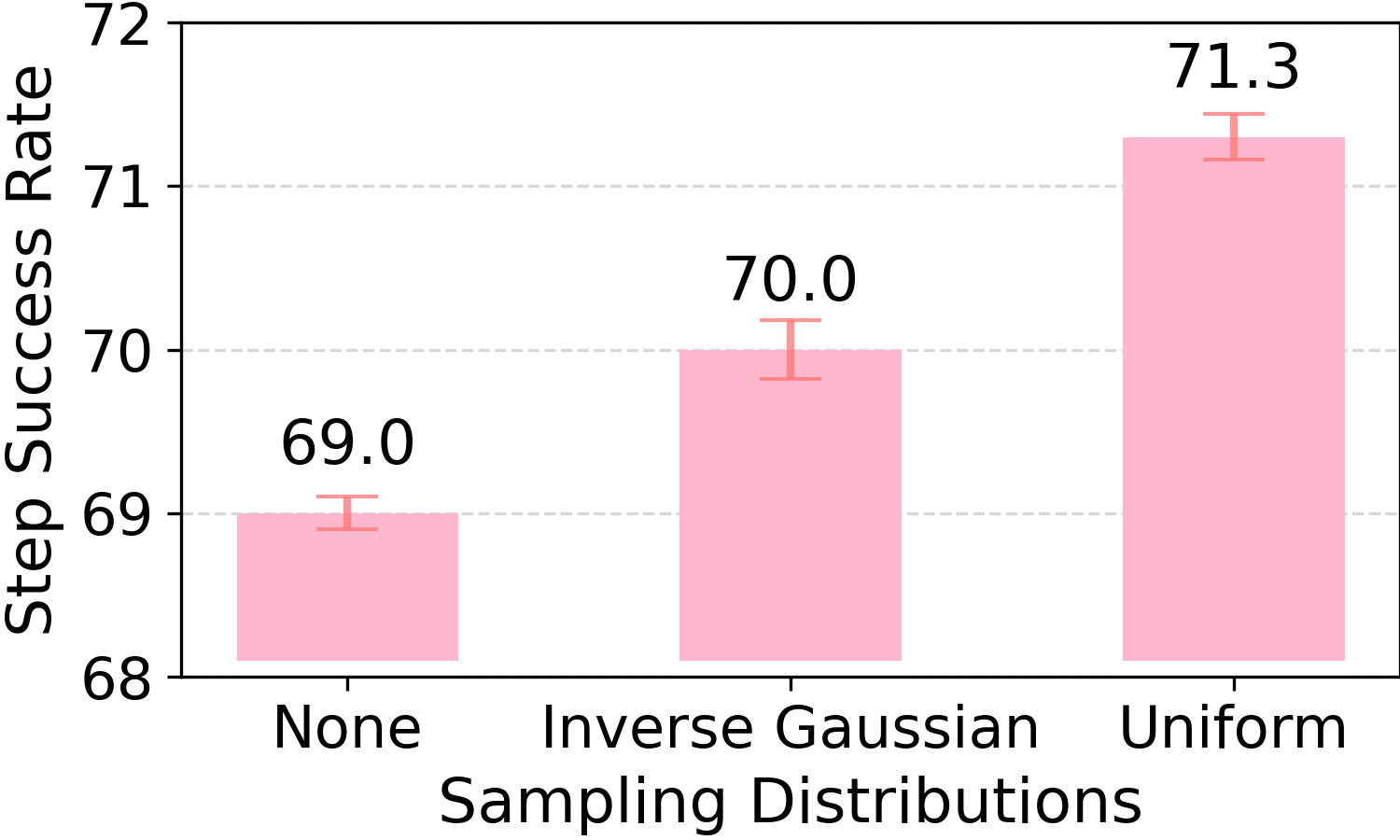}
        \caption*{(b)}
        % \label{tab:mask_dists}
    \end{minipage}
    \caption{(a): The analysis of masking region ($h,w \sim U(a,b)$) in masking-based element pruning module. (b): The analysis of masking strategy based on different sampling distributions.}
    \label{fig:mask_ana}
    \vspace{-5mm}
\end{figure}

\textbf{The analysis of masking region choice.} In Figure \ref{fig:mask_ana} (a), we analyze the selection of parameters for the size of mask region, where both $h$ and $w$ are independently sampled from $U(a,b)$. We fix the masking probability to 0.5 and sample the center point $p_c$ from a uniform distribution. Under different settings of the $a,b$ parameters, our method consistently improves without significant performance fluctuations, proving the effectiveness and insensitivity of the approach. When parameters a and b are selected from the range (0.5, 0.7), the corresponding mask will cover 25\% to 50\% of the image region. Even with such a large masked area, it still effectively enhances the model's learning, indicating that unrelated elements occupy a significant portion of the screenshot.

% \begin{table}[h]\small
%     \centering
%     \setlength{\abovecaptionskip}{1mm}
%     \begin{tabular}{c|ccc}
%     \toprule[1.2pt]
%         Sampling & \multirow{2}{*}{None} & Inverse & \multirow{2}{*}{Uniform} \\ 
%         Distribution& & Gaussian &  \\
%     \midrule
%        Step SR & 69.0$\pm0.09$ & 70.0$\pm0.19$ & \textbf{71.3}$\mathbf{\pm0.14}$  \\
%     \bottomrule[1.2pt]
%     \end{tabular}
%     \caption{The analysis of masking strategy based on different sampling distributions.}
%     \label{tab:mask_dists}
%     \vspace{-3mm}
% \end{table}

\textbf{The analysis of different masking distributions.} In Figure \ref{fig:mask_ana} (b), we analyze the impact of using uniform and inverse gaussian distributions for sampling the center point of masking regions. We set the masking probability $p=0.5$ and the size $h,w \sim U(0.5,0.7)$. The design of the Inverse Gaussian distribution is based on an intuitive prior: elements closer to the current operation point are more relevant to the task and should be masked with a smaller probability. Inverse Gaussian distribution achieves a performance improvement of 1\% compared to the non-mask variants, but it still underperforms the Uniform distribution. Notably, we also report the standard deviations, which are no more than 0.2\%, indicating the robustness of our masking strategy. Experimental results suggest that the element distribution is more complex than initially assumed. This could be due to the intricate relationships between different elements in the GUI design, where the applications have highly diverse operational logic and layouts \cite{lowgren2004}. Therefore, adopting a uniform distribution could be a solid choice, but a better solution remains to be explored.

\begin{table*}[h]\footnotesize
    \centering
    \setlength{\abovecaptionskip}{0mm}
    \begin{minipage}[t]{0.45\textwidth}
    \vspace{0pt}
        \centering
        \begin{tabular}{c|llll}
        \toprule[1.2pt]
           $k$ & $1$ & $3$ & $6$ & $12$ \\
        \midrule
           Step SR  & 66.5 & \textbf{67.3} & 66.8 & 67.4 \\
           FLOPs (T) & 8.49$_{\textcolor{teal}{\downarrow\textbf{\scriptsize29\%}}}$ & \textbf{8.71}$\mathbf{_{\textcolor{teal}{\downarrow\textbf{\scriptsize27\%}}}}$ & 9.03$_{\textcolor{teal}{\downarrow\textbf{\scriptsize24\%}}}$ & 9.67$_{\textcolor{teal}{\downarrow\textbf{\scriptsize19\%}}}$ \\
        \bottomrule[1.2pt]
        \end{tabular}
        \caption{The analysis of drop layer $k$ in history compression module without consistency guidance. The uncompressed FLOPs and performance are 11.9 and 69.0.}
        \label{tab:compression_ana}
    \end{minipage}
    \hfill
    \begin{minipage}[t]{0.5\textwidth}
    \vspace{0pt}
        \begin{tabular}{c|cccccc}
        \toprule[1.2pt]
            \multirow{2}{*}{Method} & \multirow{2}{*}{No} & Token & \multirow{2}{*}{Victor} & \multirow{2}{*}{FastV-50} & \multirow{2}{*}{FastV-0} & \multirow{2}{*}{Ours} \\
             &  & Merger  & & &  \\
        \midrule
           Step SR  & 69.0 & 68.9 & 67.6 &  66.0 & 63.8 & \textbf{68.9} \\
           FLOPs (T) & 11.90 & 9.28 & 9.28 & 10.15 & 8.71 & \textbf{8.71} \\
        \bottomrule[1.2pt]
        \end{tabular}
        \caption{Comparison to other extra-module-based (TokenMerger \cite{liu2024k}) and LLM-based (Victor\cite{wen2024b}, FastV\cite{chen2024n}) compression methods.}
        \label{tab:comp_comp}
    \end{minipage}
    % \caption{The analysis of drop layer $k$ in history compression module without consistency guidance. }
    % \label{tab:compression_ana}
\vspace{-6mm}
\end{table*}

\textbf{The analysis of drop layer choice.} In Table \ref{tab:compression_ana} we analyze the selection of the drop layer $k$ in LLM. 
It is shown that the shallow layers indeed realize a certain level of information transfer, as $k=1$ has a better performance of $66.5\%$ than the setting of utilizing only action history ($66.0\%$). After $k=3$, the step success rate reaches a saturation point (around $67.3\%$) with the minimal performance degradation. This phenomenon is also observed in recent studies \cite{chen2024n,wen2024b} on general multimodal comprehension tasks, confirming the generalization of the token compression mechanism within LLM layers across various scenarios. 

% \begin{table*}[t]
% \centering
% \caption{Main Result of xxx on GUI Odyssey.}
% \label{tb:main_guiodyssey}
% % \resizebox{1.0\textwidth}{!}{%\renewcommand\arraystretch{1.1}
% \begin{tabular}{lcccccccccc}
% \toprule[1.2pt]
% \multirow{2}{*}{Method}    & \multicolumn{2}{c}{Test-Random}  & \multicolumn{2}{c}{Test-Task}  & \multicolumn{2}{c}{Test-Device} & \multicolumn{2}{c}{Test-App} & \multicolumn{2}{c}{Overall} \\ 
%  & SSR & SR & SSR & SR & SSR & SR & SSR & SR & SSR & SR \\
% \hline
% Qwen-VL & 72.81 & 5.93 & 52.33 & 0.00 & 65.71 & 3.25 & 61.72 & 7.73 & 63.14 & 4.23 \\
% OdysseyAgent & 74.25 & 8.83 & 55.76 & 0.30 & 68.23 & 4.13 & 62.64 & 7.65 & 65.22 & 5.23 \\
% \rowcolor[HTML]{E7EEFE}
% Ours  &  & & & & & & & & & \\ 
% \bottomrule[1.2pt]
% \end{tabular}
% % }
% \end{table*}

\subsection{Comparison with other compression methods}

The existing compression schemes are divided into extra-module-based \cite{liu2024k,li2023m} and LLM-based compression methods \cite{chen2024n,wen2024b,ye2024a,hu2024d}. For the extra-module-based approach, we select the Token Merger \cite{liu2024k} as a representative work. It uses a token filter algorithm to select the most valuable tokens, utilizing these tokens as queries and employing cross-attention to further aggregate all the features. For the LLM-based compression method, we select Victor \cite{wen2024b} and FastV \cite{chen2024n}. These two methods both use summary tokens to compress visual information, but the former is applied in the training stage and the summary tokens are trainable to aggregate information, the latter focuses on inference acceleration by selecting the most valuable vision tokens as summary tokens based on attention scores.

For a fair comparison, we fix the number of selected (Token Merger) or summary (Victor and FastV) tokens at 64, and adopt the setting of Qwen2VL with 4 history screenshots and actions.
The Token Merger utilizes the resampler module to compress each historical image into 64 tokens. Victor introduces $\tau$ groups of 64 summary tokens for $\tau$ history screenshots. After the third layer, only the summary tokens are retained and the historical vision tokens are discarded. FastV is applied to the trained Qwen2VL model, where ``-0" means 0\% of the historical vision tokens are selected as summary tokens during inference. ``-50" denotes 50\% of the historical vision tokens are selected.

As shown in Table \ref{tab:comp_comp}, compared to training-based methods, Token Merger and FastV, our approach has lower FLOPs and smaller performance loss, owing to the intrinsic compression mechanism of LLM and the explicit consistency constraint. In contrast, Token Merger requires the introduction of additional parameter modules, while Victor suffers from poor performance due to the lack of explicit guidance for the compression process. As an inference-only method, FastV fails to learn better compressed features. Compared to FastV-0, our approach outperforms it by 5.1\% in the step success rate metric at the same computational complexity. While FastV-50 achieves slight improvements by retaining more vision tokens, its efficiency and performance still fall short compared to our method. These comparisons demonstrate that our consistency-guided compression method can reach an optimal balance between performance and computational efficiency.

% \begin{table}[t]
%     \centering
%     \begin{tabular}{c|ccccc}
%     \toprule[1.2pt]
%        $k$ & $1$ & $3$ & $6$ & $12$ & $\infty$ \\
%     \midrule
%        Step SR  & 66.5 & 67.3 & 66.8 & 67.4 & 69.0 \\
%        FLOPs (T) & 8.49 & \textbf{8.71} & 9.03 & 9.67 & 11.90 \\
%     \bottomrule[1.2pt]
%     \end{tabular}
%     \caption{The analysis of drop layer $k$ in history compression module without consistency guidance. }
%     \label{tab:compression_ana}
% \end{table}

% \begin{table}[t]
%     \centering
%     % \begin{tabular}{c|cccccc}
%     % \toprule[1.2pt]
%     %     Method & No & Adap. Pool & Token Merger & Victor & FastV & Ours \\
%     % \midrule
%     %    Step SR  & 69.0 & 68.6 & 68.9 & 67.6 & - & 68.9 \\
%     %    FLOPs (T) & 11.90 & - & - & 8.93 & - & 8.71 \\
%     % \bottomrule[1.2pt]
%     % \end{tabular}
%     \begin{tabular}{c|cccccc}
%     \toprule[1.2pt]
%         \multirow{2}{*}{Method} & \multirow{2}{*}{No} & Token & \multirow{2}{*}{Victor} & \multirow{2}{*}{FastV-128} & \multirow{2}{*}{FastV-0} & \multirow{2}{*}{Ours} \\
%          &  & Merger  & & &  \\
%     \midrule
%        Step SR  & 69.0 & 68.9 & 67.6 &  66.0 & 63.8 & 68.9 \\
%        FLOPs (T) & 11.90 & 9.28 & 9.28 & 10.15 & 8.71 & 8.71 \\
%     \bottomrule[1.2pt]
%     \end{tabular}
%     \caption{Comparison to other extra-module-based and LLM-based compression method. Victor [c] and FastV [c] both use summary tokens to compress visual information, but the former is applied in the training stage, and the latter focuses on inference acceleration.}
%     \label{tab:comp_comp}
% \end{table}

\begin{figure}
    \centering
    \setlength{\abovecaptionskip}{0mm}
    \includegraphics[width=1.0\linewidth]{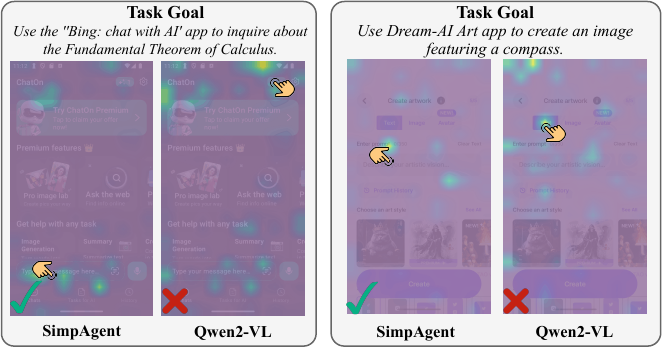}
    \caption{Illustration of some navigation steps successfully predicted by our SimpAgent, while incorrectly predicted by the baseline Qwen2-VL. The attention maps of final action tokens corresponding to the observations confirm SimpAgent can identify correct elements among distracting elements.}
    \label{fig:att_wMasking}
\vspace{-6mm}
\end{figure}

\subsection{Qualitative Analysis}
To enhance the understanding of our SimpAgent, we investigate the change of attention distribution after applying the proposed masking-based element pruning and consistency guidance.
Figure \ref{fig:att_wMasking} illustrates some navigation steps that are successfully executed by our SimpAgent, while the baseline model (fine-tuned Qwen2VL) fails to complete these steps. We present the attention maps of generated action tokens associated with the current screenshot, averaged across tokens and all attention heads.
These visualizations demonstrate that our SimpAgent can attend to the valuable elements (indicated by the peak of attention), as the masking strategy mitigates interference from unrelated elements. 

\begin{figure}
    \centering
    \setlength{\abovecaptionskip}{1mm}
    \includegraphics[width=1.0\linewidth]{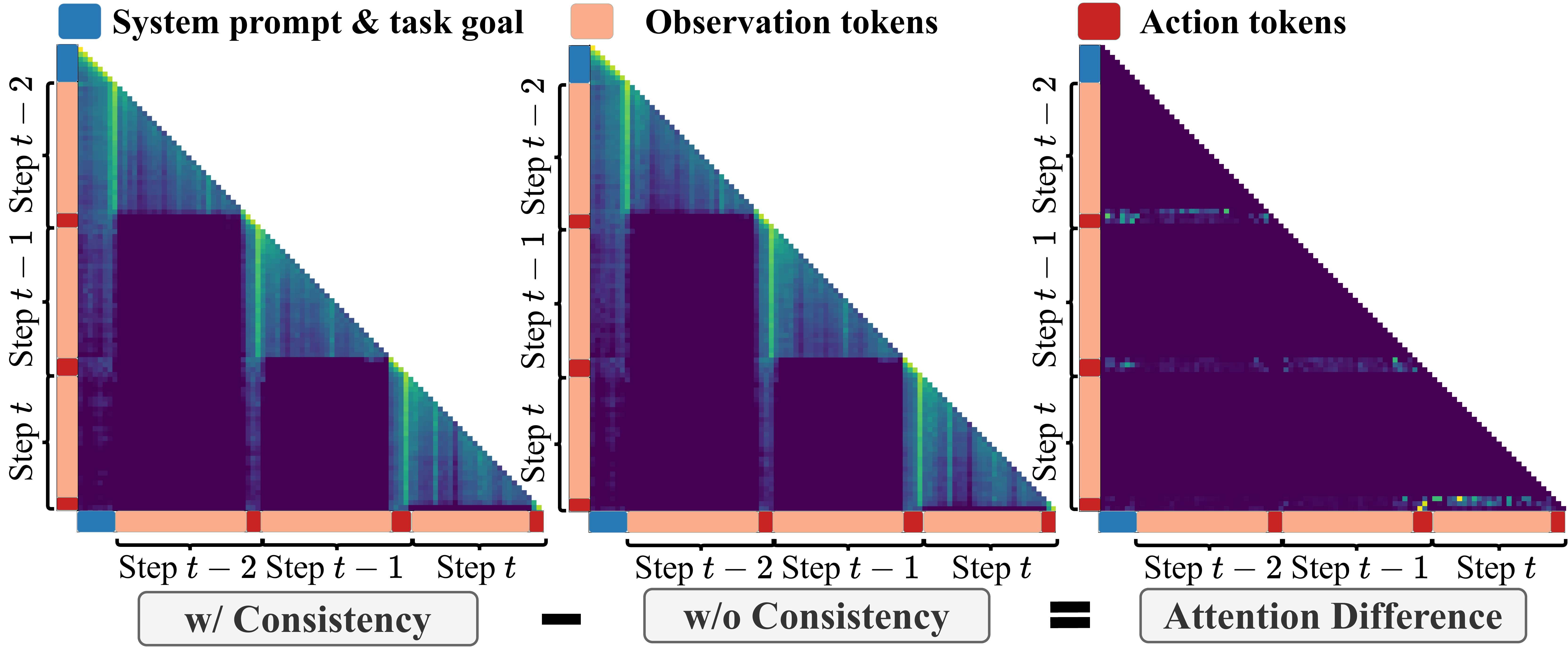}
    \caption{Illustration of attention maps in agent models w/ and w/o consistency guidance, and their attention difference map. The \textbf{attention difference map} shows that action tokens pay more attention (highlighted positions) to historical observation tokens when they act as query tokens with consistency guidance. This attention comparison demonstrates that consistency guidance can promote the information aggregation from observations to actions and facilitate the history compression.}
    \label{fig:att_wConsistency}
\vspace{-6mm}
\end{figure}

We also analyze the information flow in the agent models with and without consistency guidance. Figure \ref{fig:att_wConsistency} shows the attention maps of all tokens when the agent is trained with and without consistency guidance. To investigate the information flow of historical observations and actions, we present the attention difference map by subtracting the map without guidance from the map with consistency guidance. Note that the resulting map only highlights the difference in attention scores of action tokens to other tokens for clear comparisons. The difference map suggests that the historical action tokens pay more attention to the adjacent historical vision tokens under the setting of consistency guidance. This attention enhancement provides a clear explanation of the underlying working mechanism of the proposed consistency constraint, revealing the better information aggregation caused by this explicit supervision.

% \vspace{-3mm}
\section{Conclusion}

We introduced SimpAgent, an MLLM-based autonomous agent for GUI navigation. For building it, we devise a context-aware simplification training framework to facilitate comprehensive cognition with accurate element perception and efficient history modeling. SimpAgent is enhanced with two key components, masking-based element pruning and consistency-guided history compression. The former mitigate the interference from unrelated elements and improve the understanding of critical elements for enhanced navigation. The latter reduces the computational cost of history vision modeling, and adds the consistency guidance for better balance between performance and efficiency. SimpAgent achieves state-of-the-art performances on diverse GUI benchmarks with fewer parameters and no GUI data pre-training. These comparisons highlight SimpAgent's potential to promote the development of future GUI agents.

% \clearpage
{
    \small
    \bibliographystyle{ieeenat_fullname}
    \bibliography{references}
}

\clearpage

\appendix

\section{GUI navigation tasks}

\textbf{Android In The Wild} (AITW) \cite{rawles2023} consists of 30k instructions and 715k operation trajectories in the context of smartphone environments. However, the original train-test split carries the risk of overfitting due to overlapping instructions between the training and test sets, as well as the presence of multiple similar trajectories per instruction \cite{cheng2024}. Therefore, we adopt the instruction-wise split scheme proposed by SeeClick \cite{cheng2024} for the AITW dataset to achieve a fair evaluation. we follow the same data processing settings in SeeClick. The action space of AITW consists of 12 actions: \texttt{CLICK}, \texttt{TYPE}, \texttt{SELECT}, \texttt{SCROLL UP}, \texttt{SCROLL DOWN}, \texttt{SCROLL LEFT}, \texttt{SCROLL RIGHT}, \texttt{PRESS BACK}, \texttt{PRESS HOME}, \texttt{PRESS ENTER}, \texttt{STATUS TASK COMPLETE}, \texttt{STATUS TASK IMPOSSIBLE}.

\textbf{Mind2Web} \cite{deng2023} comprises over 2000 open-ended tasks collected from 137 real websites, originally for text-based web agents, provides only HTML observations. To support visual agents, we extracted screenshots and target bounding boxes. Since full-page captures (e.g. 1920×12000) are impractical for LVLMs, we follow the same data processing settings in SeeClick, cropping around target elements, standardizing the resolution to 1920×1080. The action space of Mind2Web consists of 2 actions:  \texttt{CLICK}, \texttt{TYPE}.

\textbf{GUI-Odyssey} \cite{lu2024a} is a comprehensive dataset for training and evaluating cross-app navigation agents, consisting of 7,735 episodes from 6 mobile devices. We follow the same data processing and evaluation settings in GUI-Odyssey. The action space of GUI-Odyssey consists of 9 actions: \texttt{CLICK}, \texttt{SCROLL}, \texttt{LONG PRESS}, \texttt{TYPE}, \texttt{COMPLETE}, \texttt{IMPOSSIBLE}, \texttt{HOME}, \texttt{BACK}, \texttt{RECENT}.

\textbf{AndroidControl} \cite{li2024a} encompasses 14,548 unique tasks across 833 Android apps, offering both high and low-level instructions to explore the level of task complexity an agent can handle. We follow the same data processing and evaluation settings in OS-Atlas \cite{wu2024c}. The action space of AndroidControl consists of 9 actions: \texttt{CLICK}, \texttt{SCROLL}, \texttt{LONG PRESS}, \texttt{TYPE}, \texttt{NAVIGATE HOME}, \texttt{NAVIGATE BACK}, \texttt{OPEN APP}, \texttt{WAIT}, \texttt{TERMINATE}.

\section{Implementation Details}
We use Qwen2VL-2B as our base model. For different datasets, we adopt the same data organization format following SeeClick \cite{cheng2024}. We parse the executed actions in JSON format, dividing them into two parts: action\_type and action\_value. Next, we introduce the format and examples of the training and testing data.

For the 4A data format, we use the same prompt in SeeClick to execute each step of the agent.
\begin{tcolorbox}[boxrule=0pt, colframe=white, sharp corners, left=1mm, right=1mm, top=0.2mm, bottom=0.2mm]
% \small
\scriptsize
{
\texttt{"<image>Please generate the next move according to the ui screenshot, instruction and previous actions.} \\
\texttt{Instruction: What's on the menu at Domino's?. Previous actions:} \\
\texttt{Step0: \{\textbackslash{}"action\_type\textbackslash{}": PRESS HOME\}.} \\
\texttt{Step1: \{\textbackslash{}"action\_type\textbackslash{}": CLICK, \textbackslash{}"click\_point\textbackslash{}": (524,865)\}.} \\
\texttt{Step2: \{\textbackslash{}"action\_type\textbackslash{}": TYPE, \textbackslash{}"typed\_text\textbackslash{}": \textbackslash{}"menu at Domino's\textbackslash{}"\}.} \\
\texttt{Step3: \{\textbackslash{}"action\_type\textbackslash{}": CLICK, \textbackslash{}"click\_point\textbackslash{}": (325,156)\}. "}
}
\end{tcolorbox}

For the 4AO data format, we use the following prompt to execute each step of the agent.
\begin{tcolorbox}[boxrule=0pt, colframe=white, sharp corners, left=1mm, right=1mm, top=0.2mm, bottom=0.2mm]
% \small
\scriptsize
{
{
\texttt{"Please generate the next move according to the instruction, previous actions, previous ui screenshot and current ui screenshot.} \\
\texttt{Instruction: What's on the menu at Domino's?.} \\
\texttt{Image\_0:<image>} \\
\texttt{Step\_0:\{\textbackslash{}"action\_type\textbackslash{}": PRESS HOME\} .} \\
\texttt{Image\_1:<image>} \\
\texttt{Step\_1:\{\textbackslash{}"action\_type\textbackslash{}": CLICK, \textbackslash{}"click\_point\textbackslash{}": (524,865)\} .} \\
\texttt{Image\_2:<image>} \\
\texttt{Step\_2:\{\textbackslash{}"action\_type\textbackslash{}": TYPE, \textbackslash{}"typed\_text\textbackslash{}": \textbackslash{}"menu at Domino's\textbackslash{}"\} .} \\
\texttt{Image\_3:<image>} \\
\texttt{Step\_3:\{\textbackslash{}"action\_type\textbackslash{}": CLICK, \textbackslash{}"click\_point\textbackslash{}": (325,156)\} .} \\
\texttt{Image\_4:<image>} "\\
}
}
\end{tcolorbox}

% 对于AITW数据集，我们follow SeeClick的4张历史图片的设置，

% 对于GUI-Odyssey，我们采用
For the AITW \cite{rawles2023} and GUI-Odyssey \cite{lu2024a} datasets, we employ four low-resolution historical images, adhering to the 4AO format proposed by SeeClick. Each image’s longest side is scaled to 512 pixels, maintaining the aspect ratio and ensuring the resolution stayed within 512x512.

For AndroidControl \cite{li2024a} and Mind2Web \cite{deng2023} datasets, we adopt the same setting in ShowUI, retaining two high-resolution historical images, with a maximum of 1280 tokens. We choose not to use four due to GPU memory constraints.

We use bfloat16 precision for training. All experiments are conducted with LoRA fine-tuning, applying a rank of 8 and an alpha value of 16 exclusively for the language model. We leverage DeepSpeed Zero-2 and utilize flash attention to accelerate training. For AITW and GUI-Odyssey, we use the same training strategy in SeeClick, with a learning rate of 3e-5 and a global batch size of 64. For AndroidControl, we use a learning rate of 3e-4 and a global batch size of 128. For Mind2Web, we use a learning rate of 3e-4 and a global batch size of 16.

% 可以补充AITW的数据集信息的图片和分析，比如element num 和 elelment size ratio
\section{The statistical Analyses of GUI Navigation Datasets}
\begin{figure}
    \centering
    \setlength{\abovecaptionskip}{0mm}
    \includegraphics[width=0.9\linewidth]{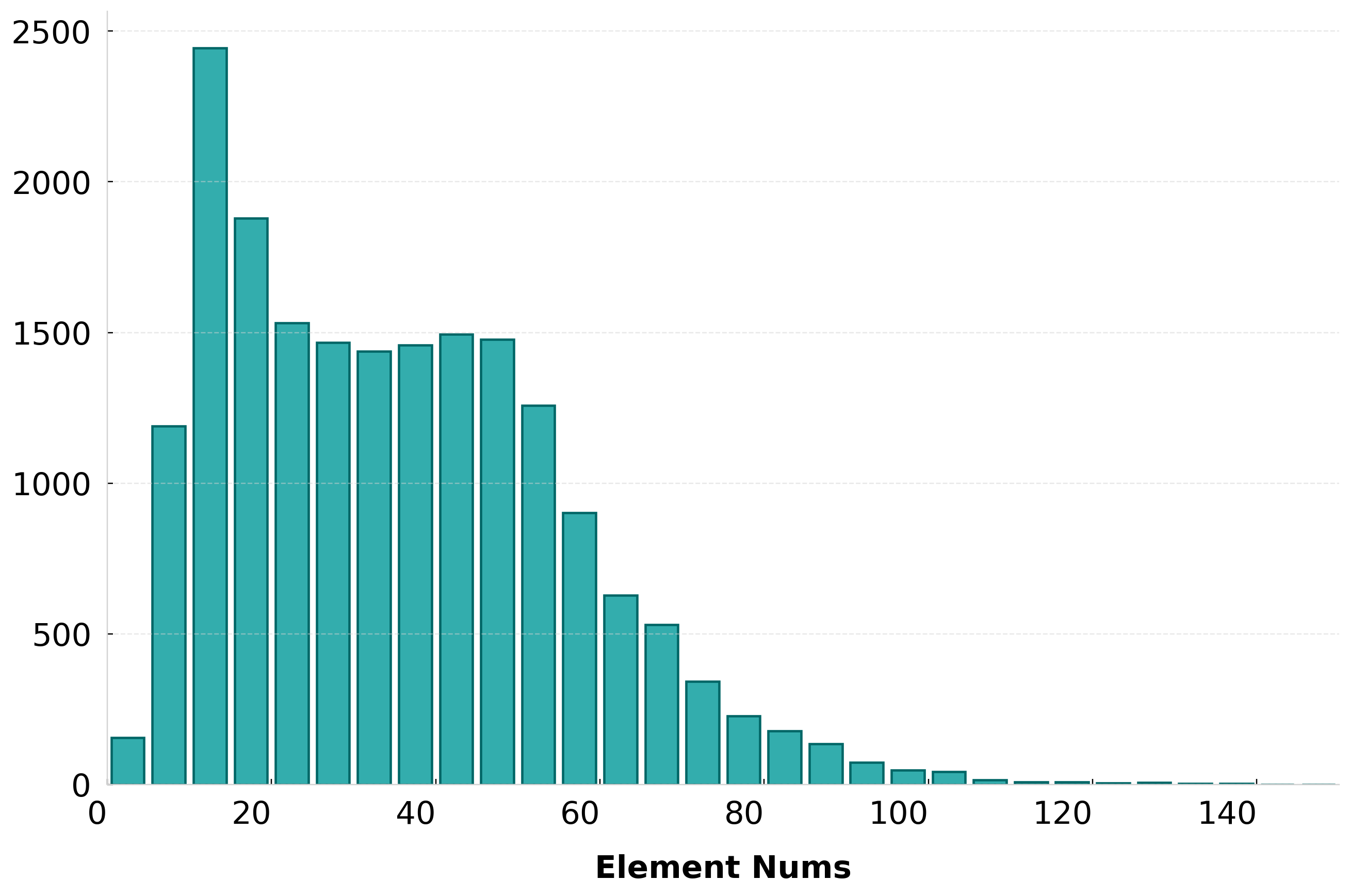}
    \caption{The distribution of element nums in AITW dataset.}
    \label{fig:element_num}
% \vspace{-3mm}
\end{figure}

In Figure \ref{fig:element_num}, We present the distribution of the element nums included in screenshots from the AITW dataset. The number of elements per screenshot typically falls within the range of 10 to 100, reflecting the characteristic of GUI scenes having a large number of visual elements.

\begin{figure}
    \centering
    \setlength{\abovecaptionskip}{0mm}
    \includegraphics[width=0.9\linewidth]{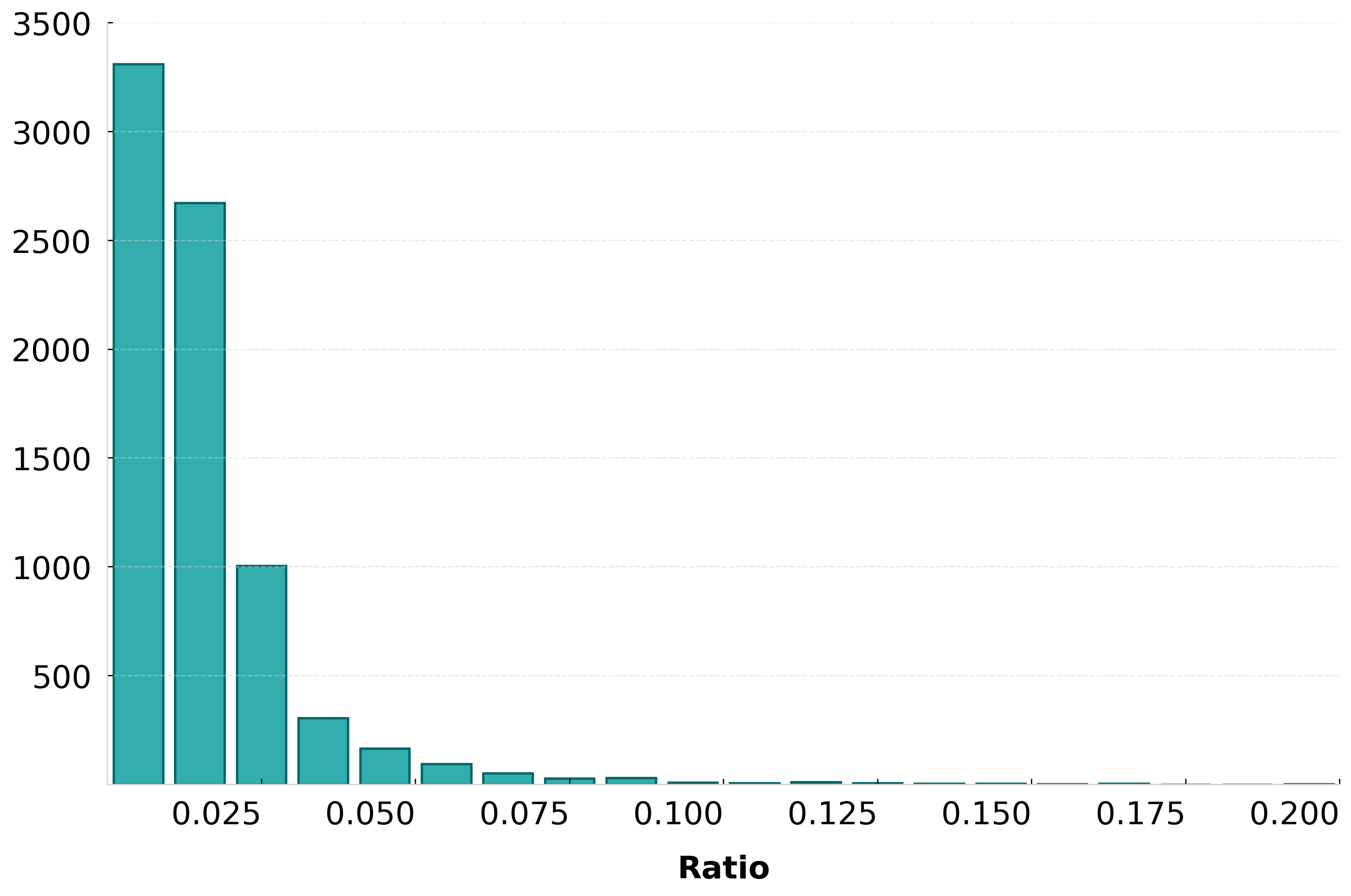}
    \caption{The distribution of target element ratio in Mind2Web dataset.}
    \label{fig:element_ratio}
% \vspace{-5mm}
\end{figure}

In Figure \ref{fig:element_ratio}, we present the distribution of the target element ratio in the Mind2Web dataset. It is evident that most elements occupy less than 3\% of the entire screen, a characteristic that contributes to the difficulty in element localization in GUI scenes.

\section{More detailed results}
In Tables \ref{tb:main_aitw}, \ref{tb:main_mind2web}, \ref{tb:main_odyssey}, we present the performances on all subsets of each navigation dataset. On AITW (6.5 steps on average), our SimpAgent significantly outperforms ShowUI, except the ``Single" subset (1.5 steps on average). We attribute it to the large-scale grounding pre-training of ShowUI, which enhances its single-horizon task completion ability. On GUI-Odyssey (15.4 steps on average), SimpAgent achieves the performance gains of 3.41\% on the ``Multi-Apps" subset (22.3 steps on average), 3.58\% on the ``Shopping" subset (18.2 steps on average), demonstrating its superior ability of accomplishing long-horizon tasks. On Mind2Web, our SimpAgent and SimpAgent-M produce new state-of-the-art performances on three out-of-domain test sets. Our SimpAgent-M is slightly inferior to SimpAgent and comparable with the baseline Qwen2VL on the ``Cross-Website" test set, and the performance gap is relatively small. We speculate that the insignificant performance difference may be due to the limited size of the ``Cross-Website" set (only 177 tasks), which is insufficient to yield reliable experimental conclusions. 

\begin{table*}[h]
\centering
\setlength{\abovecaptionskip}{1mm}
% \resizebox{1.0\textwidth}{!}{%\renewcommand\arraystretch{1.1}
\begin{tabular}{lcccccccc}
\toprule[1.2pt]
 Method & \#P.T. & \#Param. & General & Install & GoogleApps & Single & WebShopping & Overall \\
\hline
GPT-4V \cite{gpt4} & - & - & 41.7 & 42.6 & 49.8 & 72.8 & 45.7 & 50.5 \\
QwenVL \cite{bai2023} & - & 9.6B & 49.5 & 59.9 & 46.9 & 64.7 & 50.7 & 54.3 \\
SeeClick \cite{cheng2024} & 850K & 9.6B & 54.0 & 66.4 & 54.9 & 63.5 & 57.6 & 59.3 \\
% Qwen-GUI & - & 70.3 & 61.2 & 71.6 & 66.1 & \textcolor{gray}{67.3} \\
% Fuyu-GUI & - & 50.9 & 41.6 & 45.7 & 43.8 & \textcolor{gray}{43.8} \\
% MiniCPM-GUI & - & 62.3 & 46.5 & 67.3 & 67.3 & \textcolor{gray}{57.5} \\
Qwen2VL$^{*}$ \cite{wang2024n} & - & 2B & 59.4 & 71.7 & 64.2 & 71.8 & 63.0 & 66.0 \\
Qwen2VL \cite{wang2024n} & - & 2B & 61.2 & 73.5 & 72.1 & 73.8 & 64.2 & 69.0 \\
Iris \cite{ge2024} & 850K & 9.6B & 61.5 & 71.4 & 58.3 & 66.4 & 60.2 & 63.6 \\
ShowUI \cite{lin2024d} & 256K & 2B & 63.9 & 72.5 & 69.7 & \textbf{77.5} & 66.6 & 70.0 \\
\rowcolor[HTML]{E7EEFE}
SimpAgent & - & 2B & \textbf{64.3} & 75.5 & 73.7 & 75.8 & 67.0 & 71.3 \\
\rowcolor[HTML]{E7EEFE}
SimpAgent-M & - & 2B & 64.1 &  \textbf{75.8} & \textbf{74.0} & 76.2 & \textbf{67.2} & \textbf{71.5} \\ 
\bottomrule[1.2pt]
\end{tabular}
% }
\caption{Step success rate on AITW, divided by domains. ``\#P.T." and ``\#Param." denote the number of pre-training GUI datasets and parameters, respectively. ``*" means the variant utilizing only action history. ``-M" denotes only applying masking-based element pruning without inference FLOPs reduction.}
\label{tb:main_aitw}
\vspace{-3mm}
\end{table*}

\begin{table*}[h]
\centering
\setlength{\abovecaptionskip}{1mm}
% \resizebox{1.0\textwidth}{!}{%\renewcommand\arraystretch{1.1}
\begin{tabular}{lcccccccccc}
\toprule[1.2pt]
\multirow{2}{*}{Method} & \multirow{2}{*}{\#Param.} & \multicolumn{3}{c}{Cross-Task}  & \multicolumn{3}{c}{Cross-Website}  & \multicolumn{3}{c}{Cross-Domain} \\ 
\cmidrule(lr){3-5} \cmidrule(lr){6-8} \cmidrule(lr){9-11}
  & & Ele.Acc & Op.F1 & Step SR & Ele.Acc & Op.F1 & Step SR & Ele.Acc & Op.F1 & Step SR \\
\midrule
Qwen-VL \cite{bai2023} & 9.6B & 15.9 & 86.7 & 13.3 & 13.2 & 83.5 & 9.2 & 14.1 & 84.3 & 12.0 \\
CogAgent \cite{hong2023} & 18B & 22.4 & 53.0 & 17.6 & 18.4 & 42.4 & 13.4 & 20.6 & 42.0 & 15.5 \\
SeeClick \cite{cheng2024} & 9.6B & 28.3 & 87.0 & 25.5 & 21.4 & 80.6 & 16.4 & 23.2 & 84.8 & 20.8 \\
Qwen2VL$^{*}$ \cite{wang2024n} & 2B & 49.7 & 89.1 & 45.0 & 46.8 & \textbf{86.5} & 40.9 & 45.4 & 86.6 & 40.5 \\
Qwen2VL \cite{wang2024n} & 2B & 51.6 & 88.6 & 46.7 & 48.5 & 85.7 & 42.2 & 48.3 & 87.0 & 44.6 \\
Iris \cite{ge2024} & 9.6B & 33.5 & 87.1 & 32.0 & 31.2 & 82.2 & 26.2 & 32.8 & 85.1 & 28.8 \\
ShowUI \cite{lin2024d} & 2B & 39.9 & 88.6 & 37.2 & 41.6 & 83.5 & 35.1 & 39.4 & 86.8 & 35.2 \\
\rowcolor[HTML]{E7EEFE}
SimpAgent & 2B & 51.0 & 89.2 & 47.1 & \textbf{48.7} & 86.0 & \textbf{42.8} & 46.9 & 86.5 & 43.3  \\ 
\rowcolor[HTML]{E7EEFE}
SimpAgent-M & 2B & \textbf{52.4} & \textbf{89.4} & \textbf{48.7} & 48.2 & 85.8 & 42.2 & \textbf{49.0} & \textbf{88.2} & \textbf{45.0}  \\     
\bottomrule[1.2pt]
\end{tabular}
% }
\caption{Performance comparison on Mind2Web across different settings. We report element accuracy (Ele.Acc), operation F1 (Op.F1), and step success rate (Step SR). ``\#Param." denotes the number of parameters. ``*" means the variant utilizing only action history. ``-M" denotes only applying masking-based element pruning without inference FLOPs reduction.}
\label{tb:main_mind2web}
% \vspace{-5mm}
\end{table*}

\begin{table*}[h]
\centering
\setlength{\abovecaptionskip}{1mm}
\begin{tabular}{lccccccccc}
\toprule[1.2pt]
Method & \#P.T. & \#Param.  & Tool & Information & Shopping & Media & Social & Multi-Apps & Overall \\
\hline
GPT-4V \cite{gpt4} & - & - & 23.49 & 20.16 & 19.15 & 16.92 & 13.83 & 19.02 & 18.76 \\
GPT-4o \cite{hurst2024gpt} & - & - & 20.81 & 16.28 & 31.91 & 15.38 & 21.28 & 16.67 & 20.39 \\
% GeminiProVision & - & 4.03 & 2.33 & 4.26 & 1.54 & 3.19 & 4.29 & 3.27 \\
% InternVL-1.5 &  & 3.36 & 1.55 & 4.26 & 1.54 & 3.19 & 4.91 & 3.14 \\
CogAgent \cite{hong2023} & 140M & 18B & 15.66 & 10.74 & 9.15 & 11.66 & 13.08 & 10.73 & 11.84 \\
QwenVL \cite{bai2023} & - & 9.6B & 83.11 & 65.70 & 62.43 & 76.38 & 76.12 & 73.10 & 72.81 \\
Qwen2VL$^{*}$ \cite{wang2024n} & - & 2B & 80.35 & 62.42 & 59.52 & 71.11 & 72.75 & 67.91 & 69.01 \\
Qwen2VL \cite{wang2024n} & - & 2B & 83.71 & 67.44 & 64.79 & 77.19 & 79.08 & 76.97 & 74.86 \\
OdysseyAgent \cite{lu2024a} & - & 9.6B & 85.16 & 68.53 & 62.87 & 76.49 & 77.61 & 74.83 & 74.25 \\
\rowcolor[HTML]{E7EEFE}
SimpAgent & - & 2B & 84.91 & 69.89 & 66.45 & 78.05 & 79.21 & 78.24 & 76.02\\ 
\rowcolor[HTML]{E7EEFE}
SimpAgent-M & - & 2B & \textbf{86.57} & \textbf{70.08} & \textbf{67.65} & \textbf{79.78} & \textbf{81.04} & \textbf{79.21} & \textbf{77.39}\\ 
\bottomrule[1.2pt]
\end{tabular}
\caption{Step success rate on GUI-Odyssey Test-Random split. ``\#P.T." and ``\#Param." denote the number of pre-training GUI datasets and parameters, respectively. ``*" means the variant utilizing only action history. ``-M" denotes only applying masking-based element pruning without inference FLOPs reduction.}
\label{tb:main_odyssey}
% \vspace{-5mm}
\end{table*}

% 完整的mask element pruning图片和trajectory analysis
\section{Case Study}
\begin{figure*}
    \centering
    \setlength{\abovecaptionskip}{0mm}
    \setlength{\belowcaptionskip}{0mm}
    \includegraphics[width=1.0\textwidth]{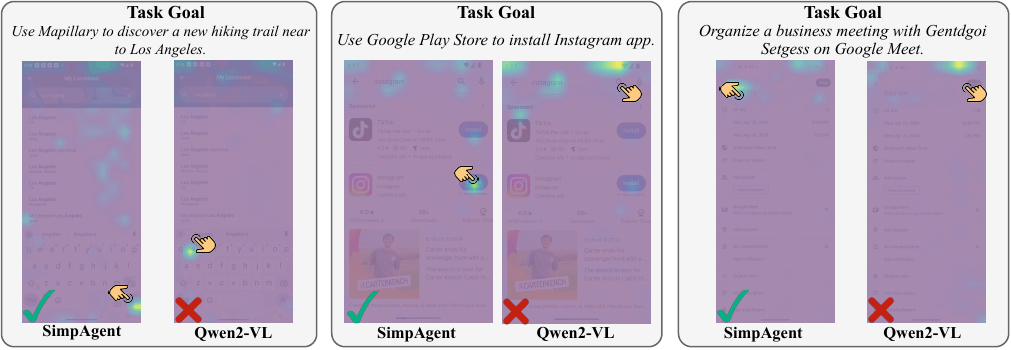}
    \caption{Illustration of navigation steps in the GUI-Odyssey dataset. SimpAgent distinguishes the correct element among various confusing elements. This demonstrates the effectiveness of our proposed Masking-based Element Pruning method.}
    \label{fig:Attn_vis}
% \vspace{-7mm}
\end{figure*}

\begin{figure*}
    \centering
    \setlength{\abovecaptionskip}{0mm}
    \setlength{\belowcaptionskip}{0mm}
    \includegraphics[width=1.0\textwidth]{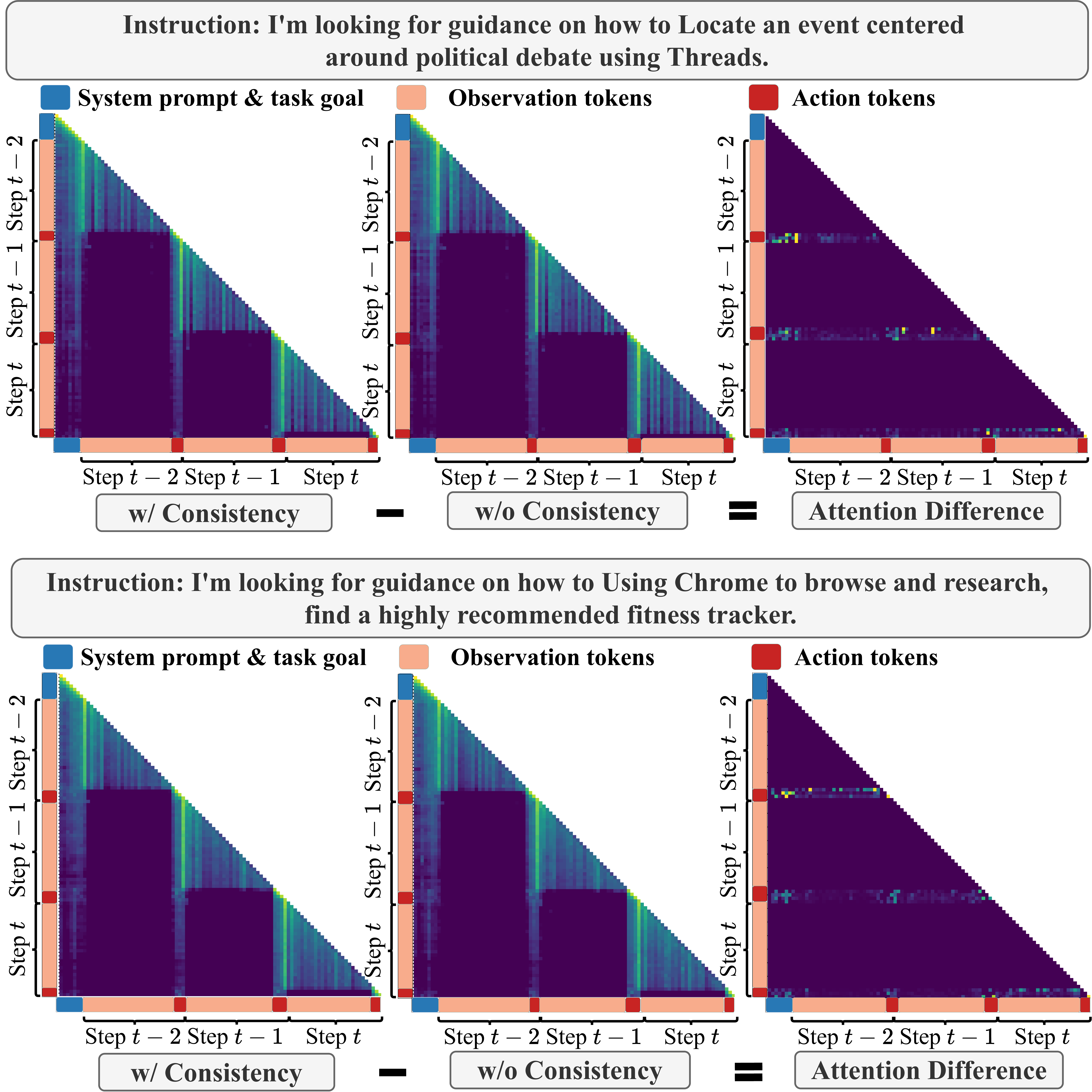}
    \caption{Illustration of attention maps in agent models w/ and w/o consistency guidance, and their attention difference map. The \textbf{attention difference map} shows that action tokens pay more attention (highlighted positions) to historical observation tokens when they act as query tokens with consistency guidance. This attention comparison demonstrates that consistency guidance can promote the information aggregation from observations to actions and facilitate the history compression.}
    \label{fig:Attn_Case}
% \vspace{-7mm}
\end{figure*}

\begin{figure*}
    \centering
    \setlength{\abovecaptionskip}{0mm}
    \setlength{\belowcaptionskip}{0mm}
    \includegraphics[width=1.0\textwidth]{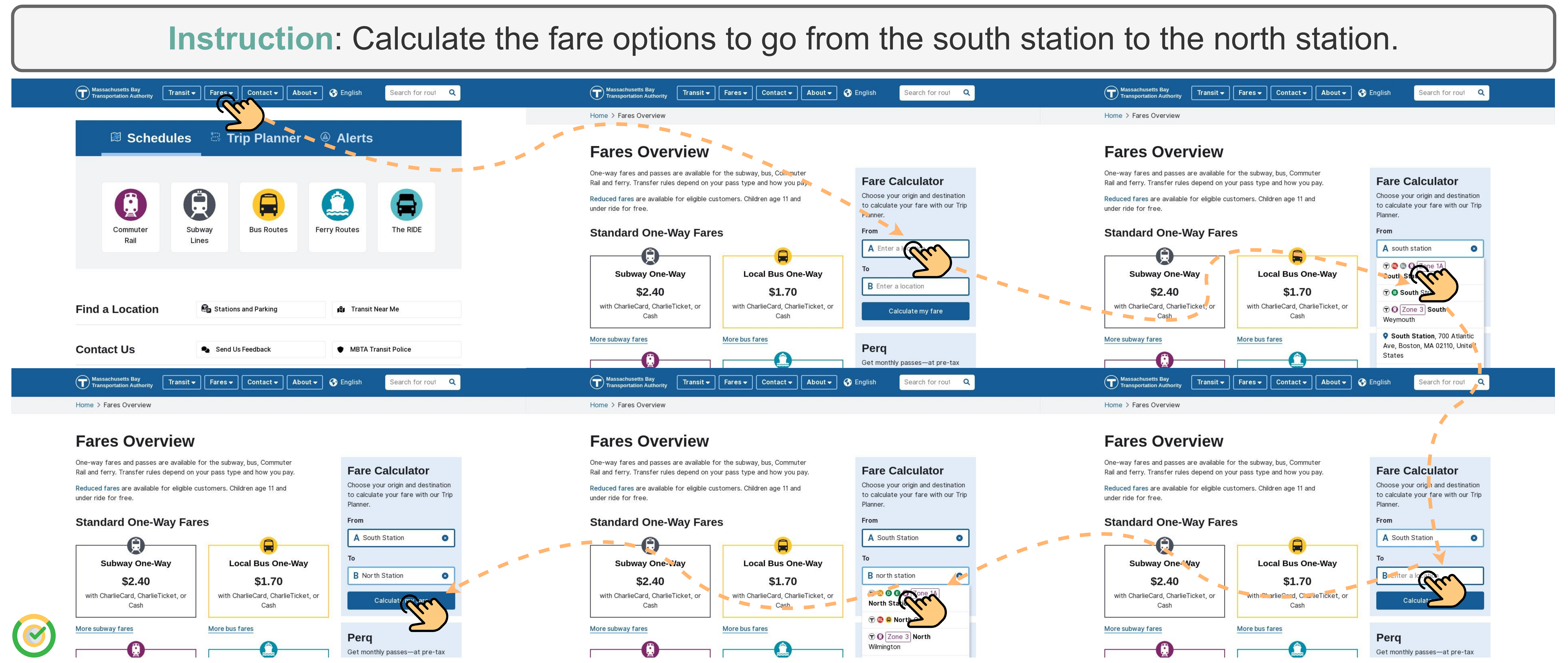}
    \caption{The real-world application of SimpAgent when adapted to downstream GUI navigation tasks.}
    \label{fig:episode_case1}
% \vspace{-7mm}
\end{figure*}

\begin{figure*}
    \centering
    \setlength{\abovecaptionskip}{0mm}
    \setlength{\belowcaptionskip}{0mm}
    \includegraphics[width=1.0\textwidth]{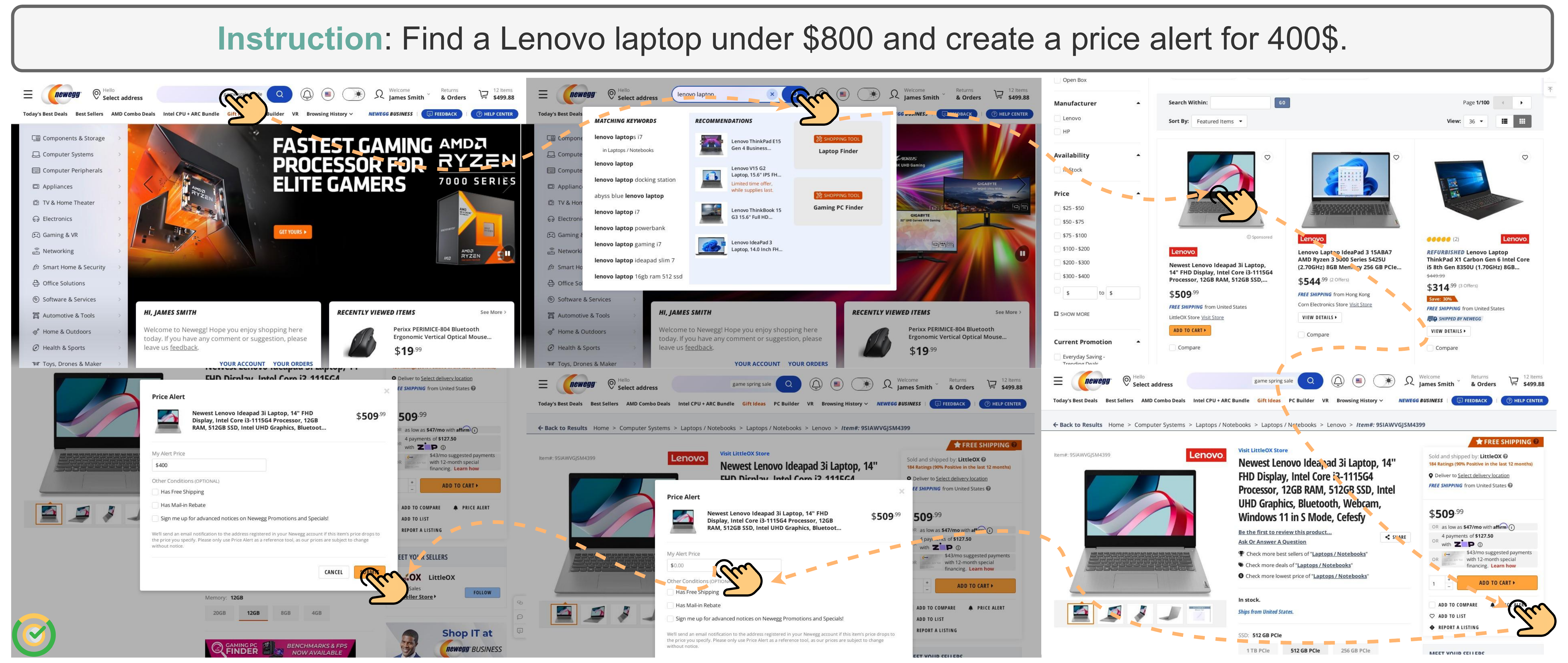}
    \caption{The real-world application of SimpAgent when adapted to downstream GUI navigation tasks.}
    \label{fig:episode_case2}
% \vspace{-7mm}
\end{figure*}

In Figure \ref{fig:Attn_vis}, we present more navigation tasks. The baseline model (fine-tuned Qwen2-VL) is unable to complete these steps. In contrast, the visualizations show that our SimpAgent can effectively reduce interference and correctly identify relevant elements while ignoring irrelevant ones. In Figure \ref{fig:Attn_Case}, we present more cases about the attention difference between w/ and w/o consistency guidance. These visualizations demonstrate that the historical action tokens pay more attention to the adjacent historical vision tokens under the setting of consistency guidance, facilitating a better information aggregation. In Figure \ref{fig:episode_case1} and \ref{fig:episode_case2}, we present the complete task execution trajectory of SimpAgent.

% \begin{figure*}
%     \centering
%     \setlength{\abovecaptionskip}{0mm}
%     \setlength{\belowcaptionskip}{0mm}
%     \includegraphics[width=1.0\textwidth]{figs/case_study/full_case.pdf}
%     \caption{Illustration of navigation steps in the GUI-Odyssey dataset. SimpAgent distinguishes the correct element among various confusing elements. This demonstrates the effectiveness of our proposed Masking-based Element Pruning method.}
%     \label{fig:full_case}
% % \vspace{-7mm}
% \end{figure*}

% \begin{figure*}
%     \centering
%     \setlength{\abovecaptionskip}{0mm}
%     \setlength{\belowcaptionskip}{0mm}
%     \includegraphics[width=1.0\textwidth]{figs/Attn_Case2.pdf}
%     \caption{The real-world application of SimpAgent when adapted to downstream GUI navigation tasks.}
%     \label{fig:Attn_Case2}
% % \vspace{-7mm}
% \end{figure*}

\end{document}